# DeepSOFA: A Continuous Acuity Score for Critically Ill Patients using Clinically Interpretable Deep Learning

**Benjamin Shickel**[1], **Tyler J. Loftus**[2], **Lasith Adhikari**[3,5], **Tezcan Ozrazgat-Baslanti**[3,5], **Azra Bihorac**[3,5,*,+], **and Parisa Rashidi**[1,4,5,+]

[1]Department of Computer and Information Science and Engineering, University of Florida, Gainesville, FL, 32611, USA.
[2]Department of Surgery, University of Florida, Gainesville, FL, 32611, USA.
[3]Department of Medicine, University of Florida, Gainesville, FL, 32611, USA.
[4]Department of Biomedical Engineering, University of Florida, Gainesville, FL, 32611, USA.
[5]Precision and Intelligent Systems in Medicine (PRISMA[P]), University of Florida, Gainesville, FL, 32611, USA.
*abihorac@ufl.edu
+these authors contributed equally to this work

## ABSTRACT

Traditional methods for assessing illness severity and predicting in-hospital mortality among critically ill patients require time-consuming, error-prone calculations using static variable thresholds. These methods do not capitalize on the emerging availability of streaming electronic health record data or capture time-sensitive individual physiological patterns, a critical task in the intensive care unit. We propose a novel acuity score framework (DeepSOFA) that leverages temporal measurements and interpretable deep learning models to assess illness severity at any point during an ICU stay. We compare DeepSOFA with SOFA (Sequential Organ Failure Assessment) baseline models using the same model inputs and find that at any point during an ICU admission, DeepSOFA yields significantly more accurate predictions of in-hospital mortality. A DeepSOFA model developed in a public database and validated in a single institutional cohort had a mean AUC for the entire ICU stay of 0.90 (95% CI 0.90-0.91) compared with baseline SOFA models with mean AUC 0.79 (95% CI 0.79-0.80) and 0.85 (95% CI 0.85-0.86). Deep models are well-suited to identify ICU patients in need of life-saving interventions prior to the occurrence of an unexpected adverse event and inform shared decision-making processes among patients, providers, and families regarding goals of care and optimal resource utilization.

## Introduction

Critically ill patients in the intensive care unit (ICU) have a life-threatening condition or the propensity to develop one at any moment. Early recognition of evolving illness severity in the ICU is invaluable. Timely and accurate illness severity assessments may identify patients in need of life-saving interventions prior to the occurrence of an unexpected adverse event and may inform shared decision-making processes among patients, providers, and families regarding goals of care and optimal resource utilization.

One of the most commonly used tools for assessing ICU patient acuity is the Sequential Organ Failure Assessment (SOFA) score [1]. SOFA considers 13 variables representing six different organ systems (cardiovascular, respiratory, nervous, liver, coagulation, and renal) and uses their worst measurements over a given interval (typically 24 hours) in conjunction with static value thresholds to assign numerical scores for each component. The sum of these component scores yields the overall SOFA score, which can be used to assess illness severity and predict mortality [2–4]. Although SOFA provides a reasonably accurate assessment of a patient's overall condition and mortality risk, its accuracy is hindered by fixed cutoff points for each component score, and SOFA variables are often infrequent or missing in electronic health records. In particular, Glasgow Coma Scale scores and measurements of serum bilirubin and partial pressure of arterial oxygen are often sparse. Badawi et al. [5] performed retrospective hourly recalculations of several acuity scores for ICU patients at 208 hospitals in the Philips eICU Research Institute database, reporting that hourly SOFA scores predicted ICU mortality with mean area under the receiver operating characteristic curve (AUC) of 0.86. Although hourly acuity score calculations may provide advantages despite the potentially confounding impact of transient and self-limited fluctuations in real-time data [6], they are only feasible if implemented as an autonomous real-time process.

The availability of temporal trends and high-fidelity physiologic measurements in the ICU offers the opportunity to apply computational approaches beyond existing conventional models [7–9]. Our primary aim was to develop an acuity score framework that encompasses the full scope of a patient's physiologic measurements over time to generate dynamic in-hospital mortality predictions. Our solution uses deep learning, a branch of machine learning that encompasses models and architectures that learn optimal features from the data itself, capturing increasingly complex representations of raw data by combining layers of nonlinear data transformations [10,11]. Deep learning models automatically discover latent patterns and form high-level representations from large amounts of raw data without the need for manual feature extraction based on *a priori* domain knowledge or practitioner intuition, which is time-consuming and error-prone. Deep learning has revolutionized natural language processing, speech recognition, and computer vision, and is gaining momentum within healthcare [12]. Computer vision has been used to identify diabetic retinopathy [13] and recognize skin cancer with accuracy similar to that of a board-certified dermatologist [14]. Deep models have also been used to predict pain responses [15], the onset of heart failure [16], and ICU mortality [17].

Here we report the development and external validation of DeepSOFA, a deep learning model that employs a clinician-interpretable variant of recurrent neural network (RNN) to analyze multivariate temporal clinical data in the ICU. Experiments were performed with two independent hospital populations and were designed to be cross-institutional; we report internal and externally validated results for both hospital cohorts. Cohorts were derived from ICU admissions at the University of Florida Health Hospital and the publicly available Medical Information Mart for Intensive Care (MIMIC-III) dataset that contains records for ICU patients from the Beth Israel Deaconess Medical Center in Boston, Massachusetts [18]. We compared deep learning mortality prediction models trained on hourly measurements with baseline models using traditional SOFA score definitions and the same hourly measurements using the entirety of a patient's data stream over the same time period. Two baseline SOFA models were tested: a Bedside SOFA model using published mortality rates correlating with any given total SOFA score [2], and a Traditional SOFA model in which hourly SOFA scores are correlated with in-hospital mortality for individual patients [5]. Because deep models automatically learn the complex, nonlinear associations among input variables, we hypothesized that DeepSOFA would yield greater accuracy in predicting in-hospital mortality among ICU patients compared with traditional SOFA techniques.

## Results

### Development of DeepSOFA model

Two datasets (*UFHealth* and *MIMIC*) derived from two distinct cohorts of ICU patients from two academic medical centers, University of Florida Health (Gainesville, FL) and Beth Israel Deaconess Medical Center (Boston, MA), respectively, were used for model development and external cross validation (Table 1). The *UFHealth* cohort included 36,216 ICU admissions for 27,660 patients, and the *MIMIC* cohort included 48,948 ICU admissions for 35,993 patients. To ensure that results would be generalizable to all patients entering an ICU at any phase of a hospital admission, all ICU admissions and readmissions for all patients were analyzed. Cohorts were comparable in terms of patients characteristics and outcomes, with slightly shorter ICU stays with a median of 2.1 days ($25^{th}$-$75^{th}$ percentiles 1.2-4.1) vs. 2.9 days ($25^{th}$-$75^{th}$ percentiles 1.5-5.9), shorter time between hospital and ICU admission with a median of 0.1 hours ($25^{th}$-$75^{th}$ percentiles 0.0-24.6) vs. 7.0 hours ($25^{th}$-$75^{th}$ percentiles 1.8-21.9), longer time between ICU and hospital discharge with a median of 73.1 hours ($25^{th}$-$75^{th}$ percentiles 27.0-143.0) vs. 48.4 hours ($25^{th}$-$75^{th}$ percentiles 0.0-122.6), and greater proportion of ICU stays requiring mechanical ventilation (47.8% vs. 30.4%) for the *MIMIC* cohort compared to *UFHealth* cohort. The *MIMIC* cohort also included a greater proportion of Medical ICU admissions (38.9% vs. 25.1%) and Cardiac ICU admissions (32.4% vs. 18.3%), with fewer Surgical ICU admissions (28.7% vs. 33.0%). The DeepSOFA model was trained and internally validated with 5-fold cross validation in each cohort separately.

The internal validation demonstrated excellent AUC performance for models in each cohort (Supplementary Fig. S2, S3 online). DeepSOFA developed and validated in the *UFHealth* cohort had AUC ranging from 0.72, 95% CI 0.71-0.72 (p<0.05 compared with Bedside SOFA AUC of 0.61, 95% CI 0.60-0.62 and p<0.05 compared with Traditional SOFA AUC of 0.63, 95% CI 0.62-0.63) at one hour after ICU admission to 0.93, 95% CI 0.93-0.94 (p<0.05 compared with Bedside SOFA AUC of 0.82, 95% CI 0.81-0.83 and p<0.05 compared with Traditional SOFA AUC of 0.88, 95% CI 0.88-0.89) at time of ICU discharge. DeepSOFA developed and validated in the *MIMIC* cohort had AUC ranging from 0.67, 95% CI 0.66-0.67 (p<0.05 compared with Bedside SOFA AUC of 0.59, 95% CI 0.58-0.60 and p<0.05 compared with Traditional SOFA AUC of 0.61, 95% CI 0.61-0.62) at one hour after

ICU admission to 0.94, 95% CI 0.93-0.94 (p<0.05 compared with Bedside SOFA AUC of 0.81, 95% CI 0.80-0.82 and p<0.05 compared with Traditional SOFA AUC of 0.85, 95% CI 0.84-0.86) at time of ICU discharge.

**External validation of DeepSOFA model**
For predicting in-hospital mortality, DeepSOFA significantly outperformed traditional SOFA models in external validation cohorts regardless of which cohort was used for model development. The DeepSOFA model developed in the *MIMIC* cohort and validated in the *UFHealth* cohort had a mean AUC for the entire ICU stay of 0.90, 95% CI 0.90-0.91 (p<0.05 compared with Bedside SOFA AUC of 0.79, 95% CI 0.79-0.80 and p<0.05 compared with Traditional SOFA AUC of 0.85, 95% CI 0.85-0.86, Fig. 1A), while the DeepSOFA model developed in the *UFHealth* cohort and validated in the *MIMIC* cohort had mean AUC of 0.90, 95% CI 0.90-0.90) (p<0.05 compared to Bedside SOFA AUC of 0.78, 95% CI 0.77-0.79 and p<0.05 compared to Traditional SOFA AUC of 0.82, 95% CI 0.81-0.82, Fig. 1B).

DeepSOFA models had significantly higher AUC across all hours of ICU stays, starting at the second hour of ICU admission with AUC of 0.74 (95% CI 0.73-0.75) in the *UFHealth* cohort (compared to Bedside SOFA AUC of 0.65, 95% CI 0.64-0.66, p<0.05 and the Traditional SOFA AUC of 0.67, 95% CI 0.66-0.67, p<0.05) and AUC of 0.68 (95% CI 0.67-0.68) in the *MIMIC* cohort (compared to Bedside SOFA AUC of 0.62, 95% CI 0.62-0.63, p<0.05 and the Traditional SOFA AUC of 0.64, 95% CI 0.64-0.65, p<0.05). This advantage remained statistically significant for the remaining duration of ICU admissions. Although all models gained accuracy over time as more input data became available, DeepSOFA accuracy increased at a greater rate during the first 24 hours following ICU admission (Fig. 1A-B).

In addition to using all 14 variables in our primary DeepSOFA model, we also assessed the accuracy of six separate DeepSOFA models using only individual subsets of variables defined in each of the SOFA organ systems (Fig. 1C-D). The individual DeepSOFA component systems most predictive of mortality were central nervous system (Glasgow Coma Scale (GCS) score), respiratory (partial pressure of arterial oxygen, fraction of inspired oxygen, and mechanical ventilation status), and cardiovascular (mean arterial pressure and vasopressor administration), all relying on more frequent time series (GCS assessed every three hours, oxygen saturation and mean arterial blood pressure assessed every minute and averaged per hour).

We also examined model accuracy as a function of the predictive window being further away from the time of death or hospital discharge. As expected, all models achieved maximum AUC in the last hour of the predictive window when using data available from the entire ICU stay in both the *UFHealth* cohort (DeepSOFA AUC of 0.93, 95% CI 0.93-0.94, p<0.05 compared to Bedside SOFA AUC of 0.82, 95% CI 0.81-0.83 and p<0.05 compared to Traditional SOFA AUC of 0.88, 95% CI 0.88-0.89) and the *MIMIC* cohort (DeepSOFA AUC of 0.93, 95% CI 0.92-0.93, p<0.05 compared to Bedside SOFA AUC of 0.81, 95% CI 0.80-0.82 and p<0.05 compared to Traditional SOFA AUC of 0.85, 95% CI 0.84-0.86). Although model performance decreased slightly when prediction occurred over a longer time window, DeepSOFA retained excellent AUC above 0.87, 95% CI 0.87-0.88 in the *UFHealth* cohort and above 0.83, 95% CI 0.82-0.83 in the *MIMIC* cohort up to 100 hours away from discharge, regardless of the mortality time point of interest (Fig. 2). These findings were consistent across all development and validation cohorts.

**Usability of DeepSOFA**
To demonstrate the feasibility of clinical application, DeepSOFA and Bedside SOFA scores were applied to a single patient encounter from the *UFHealth* cohort. The patient was a 25-year-old female with cystic fibrosis who was admitted to a Medical ICU following angioembolization of the blood supply to a lung abscess and remained in the ICU for 112 hours prior to death following cardiac arrest. Figure 3A illustrates the predicted probability of death according to DeepSOFA and Bedside SOFA scores. During the second day of ICU admission, despite increased supplemental oxygen requirements and worsening chest pain (Fig. 3D-E), the patient's vital signs remained relatively stable over time (Fig. 3B), and the Bedside SOFA model continued to estimate a low probability of death (<5%). However, predicted mortality according to DeepSOFA increased during these events, and continued to increase significantly as the patient developed increased work of breathing and required procedures to decompress the stomach and place a breathing tube, estimating a 50-80% probability of death, while the Bedside SOFA model continued to estimate a 5% probability of death. The Bedside SOFA score did not reflect clinical decompensation until the time of cardiac arrest. In the final five hours before death, the Bedside SOFA model estimated a 51.5% probability of mortality, while DeepSOFA estimated a 99.6% probability of mortality.

Translating our mortality prediction task into a real-time continuous acuity score is possible by examining the predicted probability of death at each hour of a patient's ICU stay (Fig. 4, Supplementary Fig. S4 online). Given mean mortality probabilities stratified by survival status, the traditional SOFA score tended to underestimate the severity of illness, predicting relatively low chances of death for both survivors (<5%) and non-survivors (20-30%). In contrast, DeepSOFA is better equipped to quantify illness severity for non-survivors, estimating mortality probability of 60-90% among non-survivors compared with 20-40% for survivors. DeepSOFA overestimated the probability of death for survivors, but Bedside SOFA underestimated the probability of death for non-survivors by a greater margin.

**DeepSOFA Interpretability**

DeepSOFA includes added mechanisms designed to improve the human interpretability of mortality predictions (see online supplemental section *Model Details*). Our self-attention approach is designed to highlight particular time steps of the input time series that the model believes to be most important in formulating its final mortality prediction. Since DeepSOFA is focused on real-time prediction, at each new hour after ICU admission, the model learns to distribute its internal "attention" in such a way to assign more weight to time steps it deems more influential for overall prediction.

Self-attention can be visualized as a two-dimensional matrix. At each time step after ICU admission (rows), the model assigns weights to all preceding time steps (columns) in such a way that the column weights sum to 1. Figure 5 shows examples of self-attention matrices for one survivor and one non-survivor, along with raw time series aligned by hours after ICU admission. For the example survivor, the model focused on what occurred five hours after ICU admission and continued to focus on that hour for the remaining seven hours of the encounter. By consulting the raw time series, it appears that a clinically significant decrease in creatinine and clinically significant increases in urine output and GCS contributed to DeepSOFA's overall survival prediction.

Figure 3C features a modified version of self-attention, where we visualize only the diagonal of the two-dimensional matrix. This answers the question, "how important was each time step of data at the moment it was received by the model?" For the example non-survivor, we see several attention updates in the beginning and end of the ICU stay, with changes corresponding to salient changes in the clinical time series.

## Discussion

In large, heterogeneous populations of ICU patients, we have developed and externally validated a dynamic, deep learning model (DeepSOFA) that uses a time-honored illness severity score framework to predict in-hospital mortality with significantly greater accuracy than traditional methods. We also demonstrate that the deep model may be used to generate real-time prognostic data for a single patient with visual representation of model attention, indicating time periods during which model inputs made a significant impact on predictions, improving model interpretability and application. When used to predict the likelihood of death for a single patient, DeepSOFA exhibited a consistent and proportionate response to clinical events. Because DeepSOFA may be automated, it is well suited to capitalize on the emerging availability of streaming EHR data. In this regard, deep models may augment clinical decision-making by serving as an early warning system to identify patients in need of therapeutic interventions and by informing the shared decision-making processes among patients, providers, and families regarding goals of care and resource utilization by instantaneously assessing large volumes of data over time, a task which is difficult and time-consuming for clinicians.

The superior accuracy of deep models is partially attributable to their ability to learn latent structure and complex relationships from low-level data, including temporal trends in the case of recurrent neural networks. Due to their internal memory mechanisms, recurrent mortality prediction models based on sequential time series learn temporal patterns from potentially long-term dependencies in time series variables. These complex relationships are lost in traditional models, especially when applying worst value thresholds like SOFA score calculations.

Previous work has often employed multivariable regression models in predicting mortality for ICU patients. The Simplified Acute Physiology Score (SAPS) [19,20] and Mortality Probability Model (MPM) [21] have each been used to predict in-hospital mortality using data available within one hour of ICU admission. Afessa et al. [22] evaluated the accuracy of SAPS III and MPM III in predicting mortality among large cohorts (SAPS III: 16,784 patients from 281 hospitals across five continents; MPM III: 124,855 patients from 98 hospitals in the United States), and found that each had strong accuracy (SAPS III: AUC 0.85, MPM III: AUC 0.82). In the same study, the Acute Physiology and Chronic Health Evaluation (APACHE) IV [23] score was used with data from the first 24 hours of ICU admission for 110,558 patients from 45 hospitals in the United States, and achieved AUC 0.88. Although these methods have produced reasonably accurate predictions of in-hospital mortality, their accuracy is inferior to that of deep models, and their clinical application is cumbersome compared with automated models that have the capacity for integration of streaming electronic health record data.

This study was limited by using data from hospitals within a single country. Patient populations and practice patterns from UF Health and MIMIC-III may differ from that of other ICU settings, limiting the generalizability of these findings. This study is also limited by restricting the deep learning input data to SOFA components rather than the full spectrum of variables in electronic health records. Future studies should apply DeepSOFA to live streaming electronic health record data and investigate the efficacy of expanding the input variables beyond SOFA score components to include the full spectrum of variables in electronic health records.

To our knowledge, DeepSOFA is the first application of deep learning toward generating real-time patient acuity scores. Our interpretability mechanism is also a novel application of recent advances in deep learning self-attention, where sequence elements involved in the self-attention calculation are distinct hours of a patient's ICU trajectory. We utilized these attention scores to determine and visualize the severity of fundamental time series patterns and their overall effect on the resulting acuity scores, an important contribution toward the interpretability of deep learning techniques in clinical setting.

DeepSOFA models trained on time series data were more accurate than baseline SOFA models for predicting in-hospital mortality among ICU patients. Baseline SOFA models significantly underestimated the probability of death, especially among non-survivors; DeepSOFA overestimated the probability of death among survivors, albeit to a lesser degree. Magnitude of error aside, the latter is less likely to contribute to a scenario in which clinicians fail to rescue a decompensating patient, a primary concern in ICUs. DeepSOFA may be applied to individual patients, exhibiting consistent and proportionate responses to clinical events, with visual representation of the probability of death and time periods during which model inputs disproportionately contributed to predictions. These findings suggest that the SOFA score can be augmented with more nuanced and intelligent mechanisms for assessing patient acuity. Deep learning technology may be used to augment clinician decision-making by generating accurate real-time prognostic data to identify patients in need of therapeutic interventions and inform shared decision-making processes among patients, providers, and families.

## Methods

### Study Design

Using the University of Florida Health Integrated Data Repository as Honest Broker, we created a single-center longitudinal dataset (referred to as *UFHealth*) that was extracted directly from the electronic medical records derived from 84,350 patients 18 years or older at University of Florida Health during their admissions between January 1, 2012 and April 1, 2016 as well as all encounters within one-year history and one-year follow-up. All electronic health records were de-identified, except that dates of service were maintained. The dataset includes structured and unstructured clinical data, demographic information, vital signs, laboratory values, medications, diagnoses, and procedures. Among these hospital encounters, there were 33,953 distinct encounters related to 27,660 unique patients and 36,216 ICU stays in which the patient was at least 18 years old, had their ICU stay last between 4 hours and 30 days, and had at least one measurement of mean arterial pressure and either PaO2 or SpO2 (Supplementary Fig. S5 online). Identical selection criteria were applied to the publicly available MIMIC-III [18] database of 45,748 hospital encounters and 48,948 ICU stays from 35,993 patients admitted to the Beth Israel Deaconess Medical Center in Boston, Massachusetts between 2001 and 2012. The study was approved by the University of Florida Institutional Review Board 589-2016 with waiver of informed consent.

This was a retrospective study. To predict in-hospital mortality, we made predictions every hour starting when a subject first entered the ICU, with the first mortality predictions generated one hour after ICU admission and ending at the time of ICU discharge or death. Prediction modeling was limited to data accrued during ICU admission. For patients transferred out of the ICU to an intermediate care unit or hospital ward, the end-point of hospital discharge or death was assessed at the conclusion of that hospital admission. For every prediction we used all information for our selected 14 variables available in the EHR up to the time at which the prediction was made.

### Data processing

For both cohorts, all raw time series data were extracted for the 14 variables in electronic health records (mean arterial pressure, fraction of inspired oxygen, partial pressure of oxygen, mechanical ventilation status, Glasgow Coma Scale, urine output, platelet count, serum bilirubin, serum creatinine and dosing for dopamine, dobutamine, epinephrine and norepinephrine) used in the original SOFA score, as well as for blood oxygen saturation, a commonly used respiratory measurement when partial pressure of oxygen is unavailable (Table 1, Supplementary Table S2 online). Although additional variables would have likely improved mortality prediction accuracy, the deep learning models were limited to the use of SOFA input variables to facilitate direct comparison with baseline SOFA models and as a starting point for real-time continuous acuity assessment. Variable time series began at ICU admission and ended at ICU discharge or death.

Following variable extraction, measurement outliers were removed from both cohorts according to rules in Supplementary Table S2 online, which come from both expert-defined ranges and modified Z-scores. We also employed an FiO2 imputation strategy outlined in Supplementary Table S1 online and Supplementary Fig. S6 online for calculating FiO2 based on respiratory device and oxygen flow rates.

Raw time series were then resampled to an hourly frequency, taking the mean value when multiple measurements existed for the same encounter during the same one-hour window. Following resampling, gaps in the resulting time series were filled by forward-propagating previous values for vital signs and laboratory tests and substituting 0 for vasopressor rates and the use of mechanical ventilation. For all remaining missing values, including instances in which a variable was missing entirely from an admission or before the first measurement became available, clinically normal ranges defined by experts were imputed (Supplementary Table S2 online).

The primary outcome was in-hospital mortality. Discharges to hospice in which death occurred within 7 days of hospital discharge (3% of encounters in *UFHealth* and 1.1% in *MIMIC*) were treated as mortalities.

**Model Development and Analysis**

For predicting in-hospital mortality, we used a modification of a recurrent neural network (RNN) with gated recurrent units (GRU) [24], a deep learning model ideal for working with sequentially ordered temporal data. Figure 6 shows a high-level overview of our model at three increasing levels of abstraction. Background, motivation, and detailed technical specification for our model can be found in the online supplementary section *Model Details*. Briefly, the RNN internally and continuously updates its parameters based on multivariate inputs from both the current time step and previous time steps. As such, a mortality prediction incorporates patterns detected across the entirety of an ICU admission, with recognition of longer-range temporal relationships aided by the addition of GRUs.

One of the weaknesses of deep learning techniques is the inherent difficulty in understanding the relative importance of model inputs in generating the output. In the case of mortality prediction, clinicians are interested not only in the likelihood of death, but also in knowing which factors are primarily responsible for the risk of death. If such factors are modifiable, then they represent therapeutic targets. If such factors are not modifiable, then the sustained provision of life-prolonging interventions may reach futility. To improve clinical interpretability, inspired by state-of-the-art results in other deep learning domains, we modified the traditional GRU-RNN network to include a final self-attention mechanism to allow clinicians to understand why the deep network is making its predictions. At each hour during a real-time ICU stay, the model's attention mechanism focuses on salient deep representations of all previous time points, assigning relevance scores to every preceding hour that determine the magnitude of each hour's contribution to the model's overall mortality prediction. Subject to the constraint that each hour's relevance scores must sum to 1, we are able to see exactly which hours of the multivariate time series the model thinks are most important, and how sudden the shift in attention happens. An example of this interpretable attention mechanism is shown in Fig. 5 where along with a mapping back to the original input time series, the model is able to justify its mortality predictions by changes in each of the input variables.

DeepSOFA mortality predictions were compared with two baseline models using traditional SOFA scores, which were calculated at each hour using the previous 24 hours of EHR data. The mortality predictions associated with calculated SOFA scores were derived from both published mortality rate correlations with any given score [2], which we refer to as "Bedside SOFA", and to overall AUC derived from raw SOFA scores, which we refer to as "Traditional SOFA". At any hour during an ICU admission, the Bedside SOFA baseline model associated the current SOFA score with a predicted probability of mortality, as would be performed using an online calculator, in which total SOFA scores correlate with mortality ranges. The Traditional SOFA model is based on retrospective analysis that derives AUC from raw SOFA scores and outcomes, and while not suitable for real-time prediction in practice, is a reasonable and contemporary baseline and an appropriate challenger to compare with DeepSOFA. A high-level comparison between the prediction and AUC calculation for all three models used in our experiments can be found in Supplementary Table S3 online.

Our baselines are based on both current practice (Bedside SOFA) and recent retrospective methods (Traditional SOFA). Both of these baselines utilize a single feature (current SOFA score) from patient time series for making hourly predictions. As a sensitivity analysis, we also trained two additional conventional machine learning models (logistic regression, random forest) using 84 aggregate features recalculated at every hour after ICU admission, including the following for each of the 14 SOFA variables: minimum value, maximum value, mean value, standard deviation, first value, and last value. A summary of these results can be found in Supplementary Tables S4 and S5 online for internal and external validation, respectively. The SOFA baselines included in our study outperformed these additional machine learning models.

**Model Evaluations and Statistical Analysis**

Using each of *UFHealth* and *MIMIC* as primary cohorts, we performed several internal and externally-validated experiments and baseline comparisons. For each cohort, internal validation experiments consisted of training and testing the DeepSOFA model on the same data source. In this setting, we performed 5-fold cross-validation in which a model was trained on a random 80% of ICU admissions and tested on the remaining 20%, repeated for 5 non-overlapping iterations to yield a prediction trajectory for every ICU stay in the cohort. In the external validation experiments, a DeepSOFA model was trained on the entirety of one cohort and tested on the entirety of the other cohort. Baseline models did not require training and were applied to the same testing cohorts as DeepSOFA. All reported results are from the external validation experiments. Internal cross-validation results, which are more optimistic than their external counterparts, can be found in the online supplement.

Predictions involved in our experiments were performed on individual ICU encounters; as such, a single patient could have multiple ICU stays that appear as distinct prediction units. We performed a sensitivity analysis involving two variations of adjusting for patients with multiple ICU encounters in their EHR, including (1) only keeping their first ICU encounter, and (2) removing such patients entirely from the dataset. All models were retrained and tested using these modified datasets, and a summary of these prediction results can be found in Supplementary Tables S6 and S7 online for internal and external validation, respectively. Both modifications resulted in increased performance for all models.

For all models, a prediction was obtained at every hour, beginning one hour after ICU admission and ending at the time of ICU discharge or death. We assessed model discrimination by calculating area under the receiver operating characteristic curve (AUC), and calculated 95% confidence intervals using 100 bootstrapped iterations of sampling mortality prediction probabilities with replacement. At each hour, all ICU stays were included in reported results; for encounters with duration less than the current hour, the final prediction was used in AUC calculations. Figure 4 and Supplementary Fig. S4 online show the number of active ICU encounters and corresponding mortality rates by hour.

## Data availability

The *MIMIC* cohort is derived from the publicly available MIMIC-III database[18]. *UFHealth* cohort data are available from the University of Florida Institutional Data Access / Ethics Committee for researchers who meet the criteria for access to confidential data and may require additional IRB approval.

# Acknowledgements

The authors acknowledge Gigi Lipori, MBA and Paul Nickerson, MS for assistance with data retrieval.
    AB, TOB, and PR were supported by R01 GM110240 from the National Institute of General Medical Sciences. PR was supported by CAREER award, NSF-IIS 1750192, from the National Science Foundation (NSF), Division of Information and Intelligent Systems (IIS). AB and TOB were supported by Sepsis and Critical Illness Research Center Award P50 GM-111152 from the National Institute of General Medical Sciences. TOB has received grant (97071) from Clinical and Translational Science Institute, University of Florida. This work was supported in part by the NIH/NCATS Clinical and Translational Sciences Award to the University of Florida UL1 TR000064. TL was supported by a post-graduate training grant (T32 GM-008721) in burns, trauma and perioperative injury awarded by the National Institute of General Medical Sciences (NIGMS). BS was supported by internal funds at J. Crayton Pruitt Family Department of Biomedical Engineering, University of Florida. The content is solely the responsibility of the authors and does not necessarily represent the official views of the National Institutes of Health. The funders had no role in study design, data collection and analysis, decision to publish, or preparation of the manuscript. AB, BS, and TOB had full access to all the data in the study and take responsibility for the integrity of the data and the accuracy of the data analysis. The Titan X Pascal partially used for this research was donated by the NVIDIA Corporation.

# Author contributions

BS, TL, AB, LA, and TOB had full access to the data in the study and take responsibility for the integrity of the data and the accuracy of the data analysis. Study design performed by BS, TL, AB, and PR. BS and TL drafted the manuscript. Analysis was performed by BS, TL, TOB, LA, AB, and PR. Funding was obtained by AB and PR. Administrative, technical, material support was provided by AB and PR. Study supervision was performed by AB and PR. All authors contributed to the acquisition, analysis, and interpretation of data. All authors contributed to critical revision of the manuscript for important intellectual content.

# Additional information

**Competing interests**
The authors declare no competing interests.

|  | *UFHealth* (n = 36216) | *MIMIC* (n = 48948) |
|---|---|---|
| **Hospital admissions, n** | **33953** | **45748** |
| Patients, n | 27660 | 35993 |
| Female gender, n (%) | 15170 (44.7%) | 19558 (42.8%) |
| Age, median (25th, 75th) | 61.5 (49.2, 71.5) | 64.5 (52.0, 76.1) |
| Body mass index, median (25th, 75th) | 27.0 (23.1, 31.9) | 27.3 (23.7, 31.7) |
| Charlson comorbidity index, median (25th, 75th) | 2 (0, 3) | 2 (2, 2) |
| Race, n (%) | | |
|   White | 25837 (76.1%) | 32462 (71.0%) |
|   African-American | 5450 (16.1%) | 4471 (9.8%) |
|   Hispanic | 1081 (3.2%) | 1714 (3.7%) |
|   Asian | 238 (0.7%) | 1079 (2.4%) |
|   Other/Missing | 1347 (4.0%) | 6022 (13.2%) |
| Hospital LOS, days, median (25th, 75th) | 7.1 (3.9, 13.0) | 6.9 (4.1, 11.9) |
| In-hospital mortality rate, % | 10.40% | 10.80% |
| Discharged to hospice, n (%) | 1011 (3.0%) | 488 (1.1%) |
| Hospital LOS for non-survivors, days, median (25th, 75th) | 6.2 (2.4, 13.4) | 6.1 (2.2, 13.2) |
| **ICU admissions, n** | **36216** | **48948** |
| ICU stays per hospital admission, median (min, max) | 1.0 (1.0, 6.0) | 1.0 (1.0, 7.0) |
| ICU LOS, days, median (25th, 75th) | 2.9 (1.5, 5.9) | 2.1 (1.2, 4.1) |
| ICU type, n (%) | | |
|   Medical ICU | 9102 (25.1%) | 19054 (38.9%) |
|   Cardiac ICU | 6621 (18.3%) | 15844 (32.4%) |
|   Surgical ICU | 11941 (33.0%) | 14050 (28.7%) |
|   Neurological ICU | 7633 (21.1%) | 0 (0.0%) |
|   Burn ICU | 642 (1.8%) | 0 (0.0%) |
|   Pediatric ICU | 277 (0.8%) | 0 (0.0%) |
| Deaths occurring in ICU, n (% all deaths) | 2768 (71.3%) | 3873 (68.9%) |
| Hours between hospital admission and ICU admission, median (25th, 75th) | 7.0 (1.8, 21.9) | 0.1 (0.0, 24.6) |
| Hours between ICU discharge and hospital discharge/death, median (25th, 75th) | 48.4 (0.0, 122.6) | 73.1 (27.0, 143.0) |
| **Variables, n** | **14** | **14** |
| ICU stays requiring any vasopressor, n (%) | 6216 (17.2%) | 9418 (19.2%) |
| ICU stays requiring mechanical ventilation, n (%) | 11004 (30.4%) | 23410 (47.8%) |
| Urine, mL (min, max value) | (0.0, 1090.0) | (0.0, 1090.0) |
|   Frequency, hours, median (25th, 75th) | 1.0 (1.0, 1.9) | 1.0 (1.0, 1.5) |
|   Measured value, median (25th, 75th) | 100.0 (50.0, 200.0) | 85.0 (45.0, 160.0) |
|   ICU stays missing any measurement, n (%) | 1213 (3.3%) | 2080 (4.2%) |
| Bilirubin, mg/dL (min, max) | (0.1, 49.2) | (0.1, 50.0) |
|   Frequency, hours, median (25th, 75th) | 24.2 (19.0, 48.0) | 23.0 (10.8, 25.1) |
|   Measured value, median (25th, 75th) | 0.6 (0.3, 1.5) | 1.1 (0.5, 3.7) |
|   ICU stays missing any measurement, n (%) | 17278 (47.7%) | 27884 (57.0%) |
| Creatinine, mg/dL (min, max) | (0.1, 26.5) | (0.1, 29.1) |
|   Frequency, hours, median (25th, 75th) | 15.2 (6.3, 24.0) | 13.3 (8.5, 23.5) |
|   Measured value, median (25th, 75th) | 0.9 (0.7, 1.4) | 1.0 (0.7, 1.7) |
|   ICU stays missing any measurement, n (%) | 1714 (4.7%) | 942 (1.9%) |
| FiO2, % (min, max) | (21.0, 100.0) | (21.0, 100.0) |
|   Frequency, hours, median (25th, 75th) | 1.0 (1.0, 2.0) | 3.0 (2.0, 7.0) |
|   Measured value, median (25th, 75th) | 40.0 (30.0, 40.0) | 37.0 (29.0, 50.0) |
|   ICU stays missing any measurement, n (%) | 0 (0.0%) | 0 (0.0%) |
| PaO2, mmHg (min, max) | (9.0, 720.0) | (1.0, 775.0) |
|   Frequency, hours, median (25th, 75th) | 4.2 (2.6, 8.3) | 3.9 (1.8, 7.6) |
|   Measured value, median (25th, 75th) | 119.0 (87.9, 155.0) | 109.0 (83.0, 151.0) |
|   ICU stays missing any measurement, n (%) | 17257 (47.7%) | 19739 (40.3%) |
| Mean arterial pressure, mmHg (min, max) | (1.0, 300.0) | (0.4, 300.0) |
|     Frequency, hours, median (25th, 75th) | 1.0 (0.2, 1.0) | 1.0 (0.5, 1.0) |
|     Measured value, median (25th, 75th) | 79.0 (69.0, 91.0) | 77.7 (68.0, 89.0) |
|     ICU stays missing any measurement, n (%) | 0 (0.0%) | 0 (0.0%) |
| Platelet count, x 10^3/mm^3 (min, max) | (1.0, 832.0) | (5.0, 832.0) |
|     Frequency, hours, median (25th, 75th) | 17.7 (6.5, 24.0) | 18.4 (8.0, 24.0) |
|     Measured value, median (25th, 75th) | 175.0 (106.0, 257.0) | 186.0 (118.0, 273.0) |
|     ICU stays missing any measurement, n (%) | 1949 (5.4%) | 1187 (2.4%) |
| GCS, score (min, max) | (3.0, 15.0) | (3.0, 15.0) |
|     Frequency, hours, median (25th, 75th) | 1.0 (1.0, 4.0) | 4.0 (2.0, 4.0) |
|     Measured value, median (25th, 75th) | 14.0 (10.0, 15.0) | 14.0 (9.0, 15.0) |
|     ICU stays missing any measurement, n (%) | 1595 (4.4%) | 83 (0.2%) |

**Table 1.** Demographics and summary of included variables for *UFHealth* and *MIMIC* cohorts. Variable summary statistics were obtained after imputing FiO2 and removing outliers as detailed in Supplementary Table S2 online.

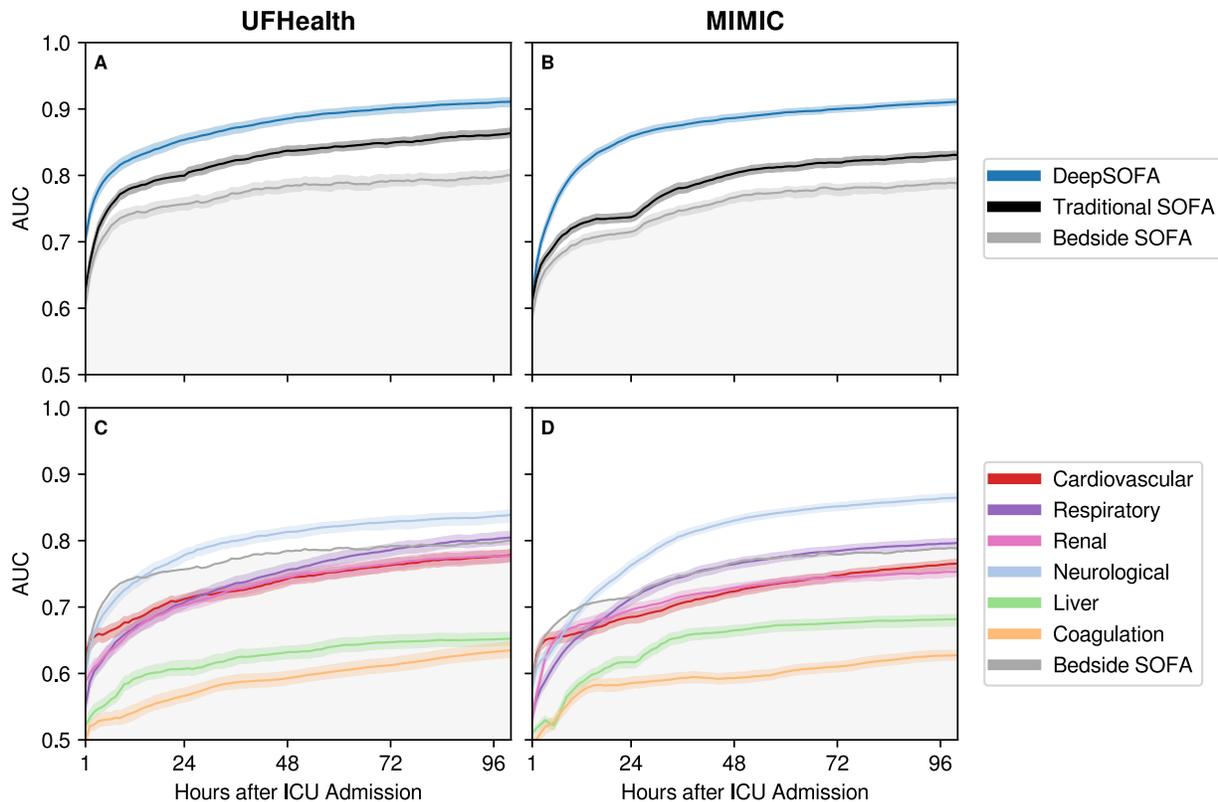

**Figure 1.** DeepSOFA performance in two external validation cohorts. (A, B) Externally validated DeepSOFA, Bedside SOFA, and Traditional SOFA score accuracy in predicting in-hospital mortality, expressed as area under the receiver operating characteristic curve (AUC) for the first 100 hours following ICU admission. (C, D) Externally validated DeepSOFA accuracy for individual models corresponding to variable sets derived from SOFA organ system classification for the first 100 hours following ICU admission. Shaded regions represent 95% confidence intervals based on 100 bootstrapped iterations. Columns specify the validation cohort. DeepSOFA model validated in *UFHealth* (A, C) was trained using *MIMIC*, and DeepSOFA model validated in *MIMIC* (B, D) was trained using *UFHealth*. SOFA: Sequential Organ Failure Assessment.

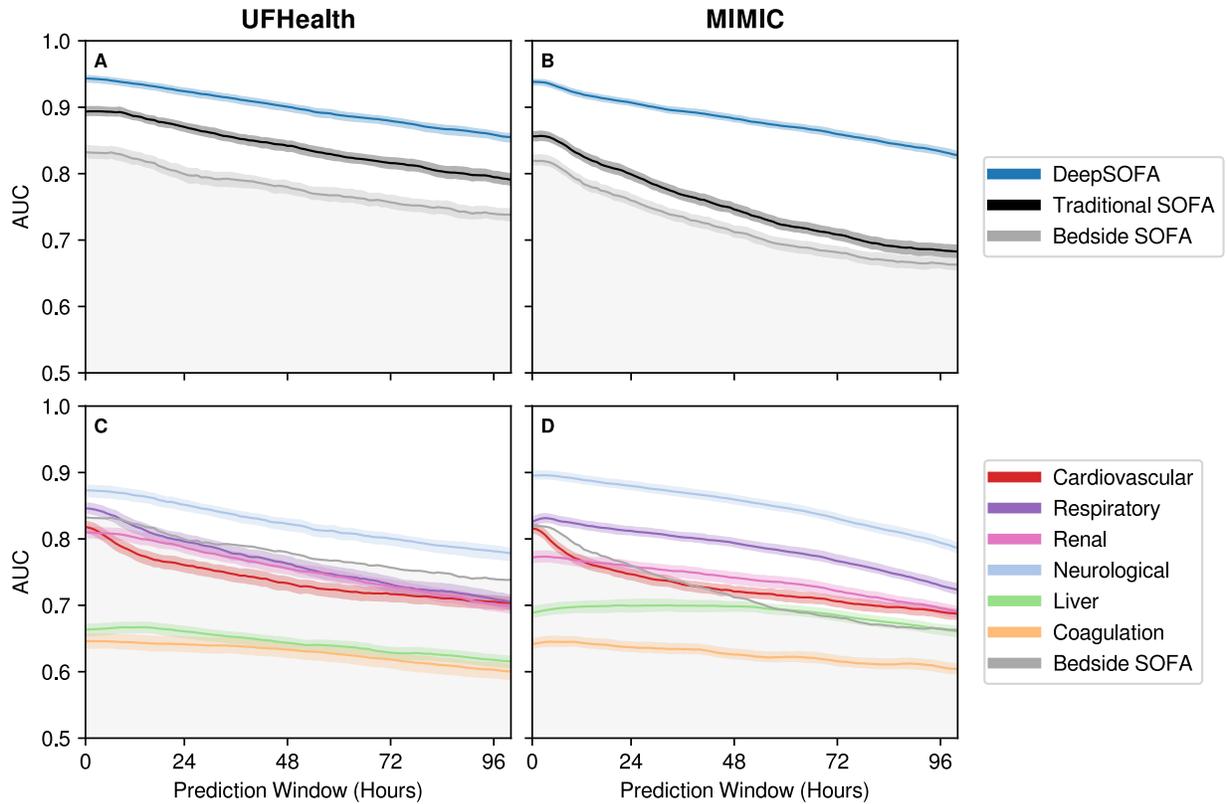

**Figure 2.** Model accuracy for prediction windows of increasing time from hospital discharge or death. (A, B) Externally validated DeepSOFA, Bedside SOFA, and Traditional SOFA score accuracy in predicting in-hospital mortality, expressed as area under the receiver operating characteristic curve (AUC) for 100 hours preceding death or hospital discharge. (C, D) Externally validated DeepSOFA accuracy for individual models corresponding to variable sets derived from SOFA organ system classification for the 100 hours preceding death or hospital discharge. Shaded regions represent 95% confidence intervals based on 100 bootstrapped iterations. Columns specify the validation cohort. DeepSOFA model validated in *UFHealth* (A, C) was trained using *MIMIC*, and DeepSOFA model validated in *MIMIC* (B, D) was trained using *UFHealth*. SOFA: Sequential Organ Failure Assessment.

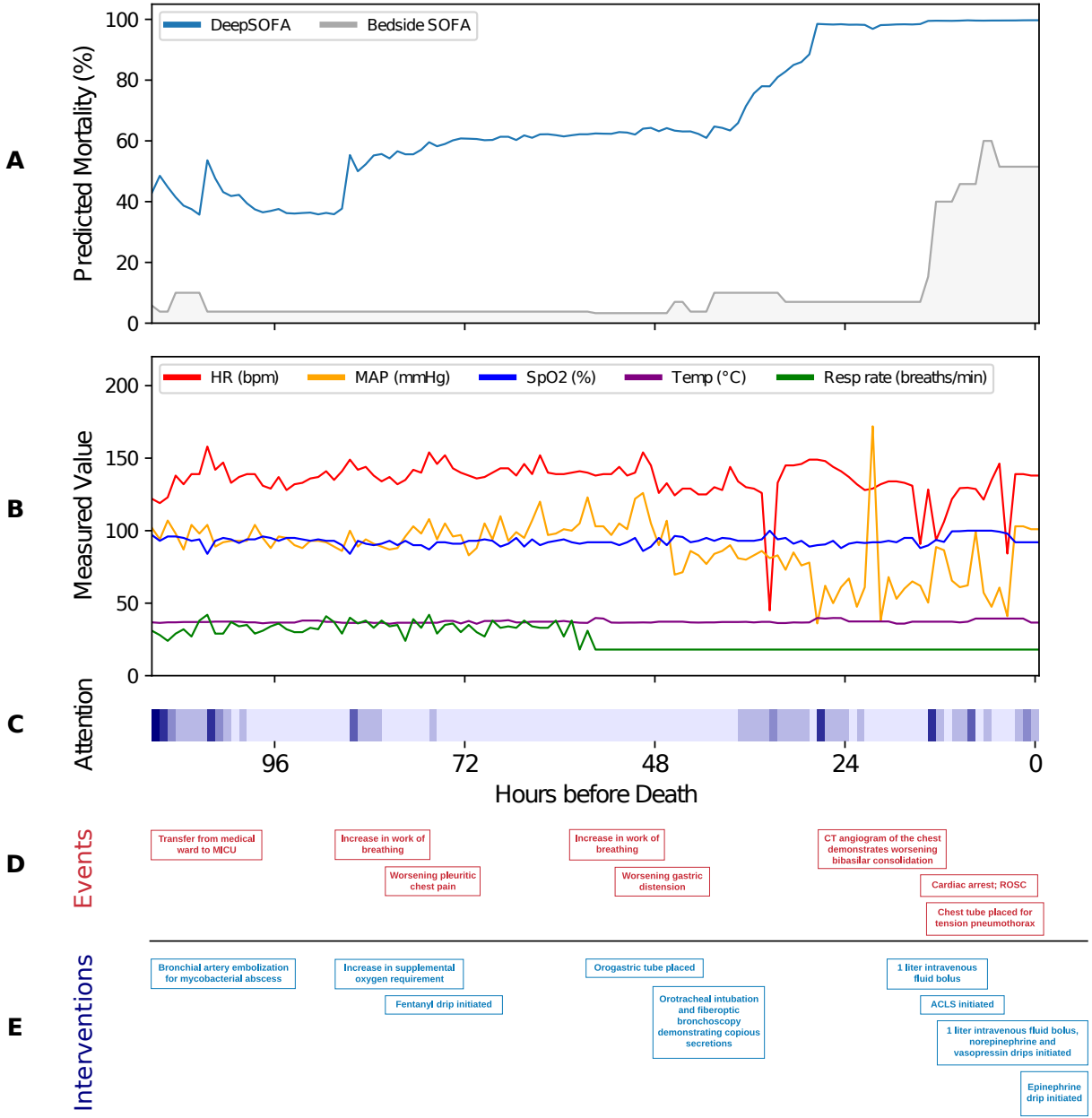

**Figure 3.** Externally validated DeepSOFA and Bedside SOFA score predicted mortality (A) for a single patient from the *UFHealth* cohort, correlated with vital signs (B), clinical events (D), and clinical interventions (E). Shown also are model self-attention weights (C) for each hour after ICU admission, where darker bars indicate increased model focus. Attention weights taken from the diagonal of full self-attention matrix and indicate how important the model believes each hour's data is, as it is encountered in real-time. SOFA: Sequential Organ Failure Assessment, SpO2: oxygen saturation, MICU: Medical Intensive Care Unit, CT: computed tomography, ACLS: Advanced Cardiac Life Support, ROSC: return of spontaneous circulation.

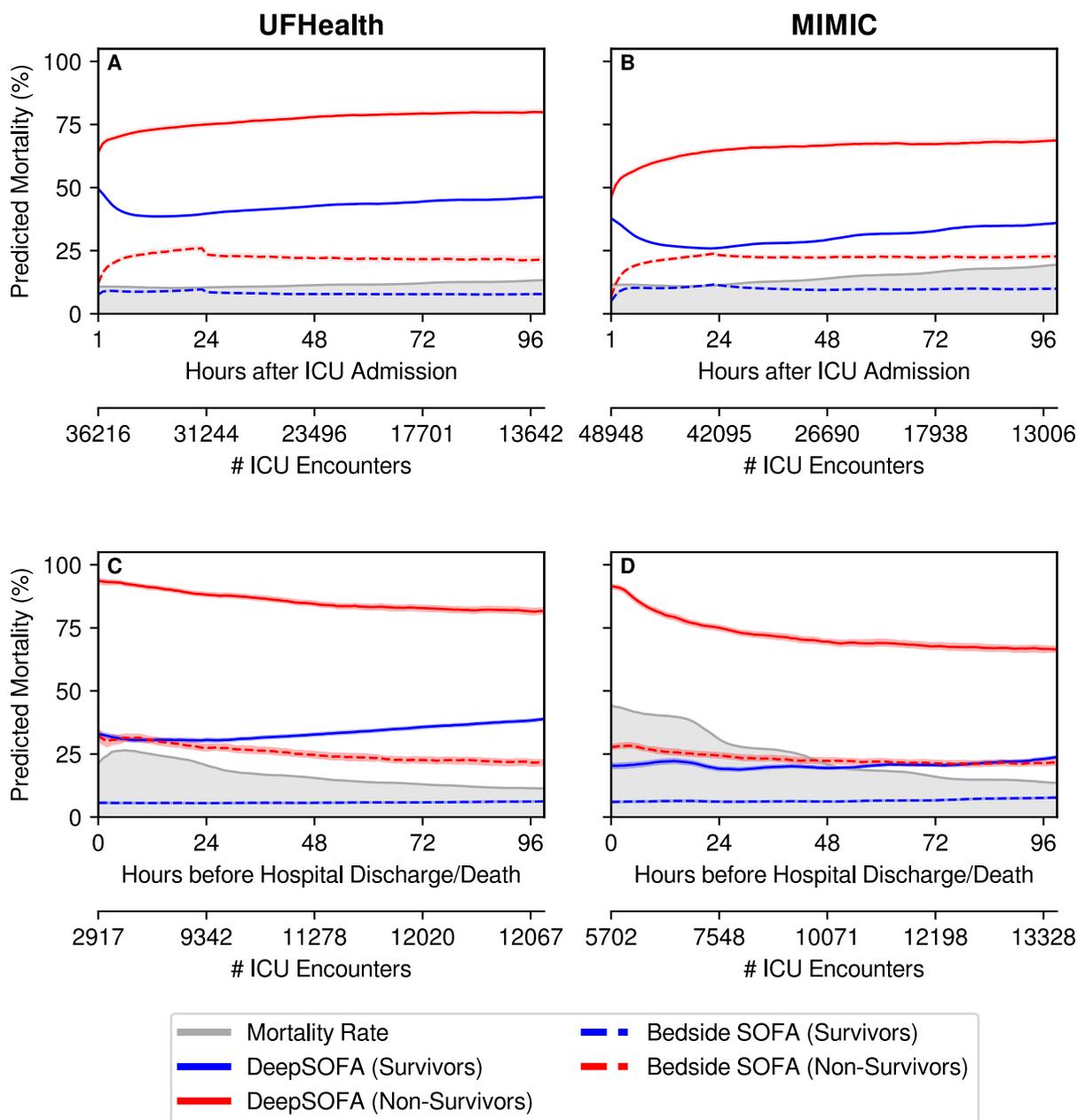

**Figure 4.** Mean predicted mortality probabilities for externally validated DeepSOFA and Bedside SOFA models stratified by outcome. Probabilities shown both for first 100 hours after ICU admission (A, B) and final 100 hours before hospital discharge or death (C, D). Number of ICU encounters at each time point shown below each panel. Shaded regions around each line represent 95% confidence intervals based on 100 bootstrapped iterations. Gray shared area denotes hourly mortality rate for active ICU encounters. Columns specify the validation cohort. DeepSOFA model validated in *UFHealth* (A, C) was trained using *MIMIC*, and DeepSOFA model validated in *MIMIC* (B, D) was trained using *UFHealth*. SOFA: Sequential Organ Failure Assessment.

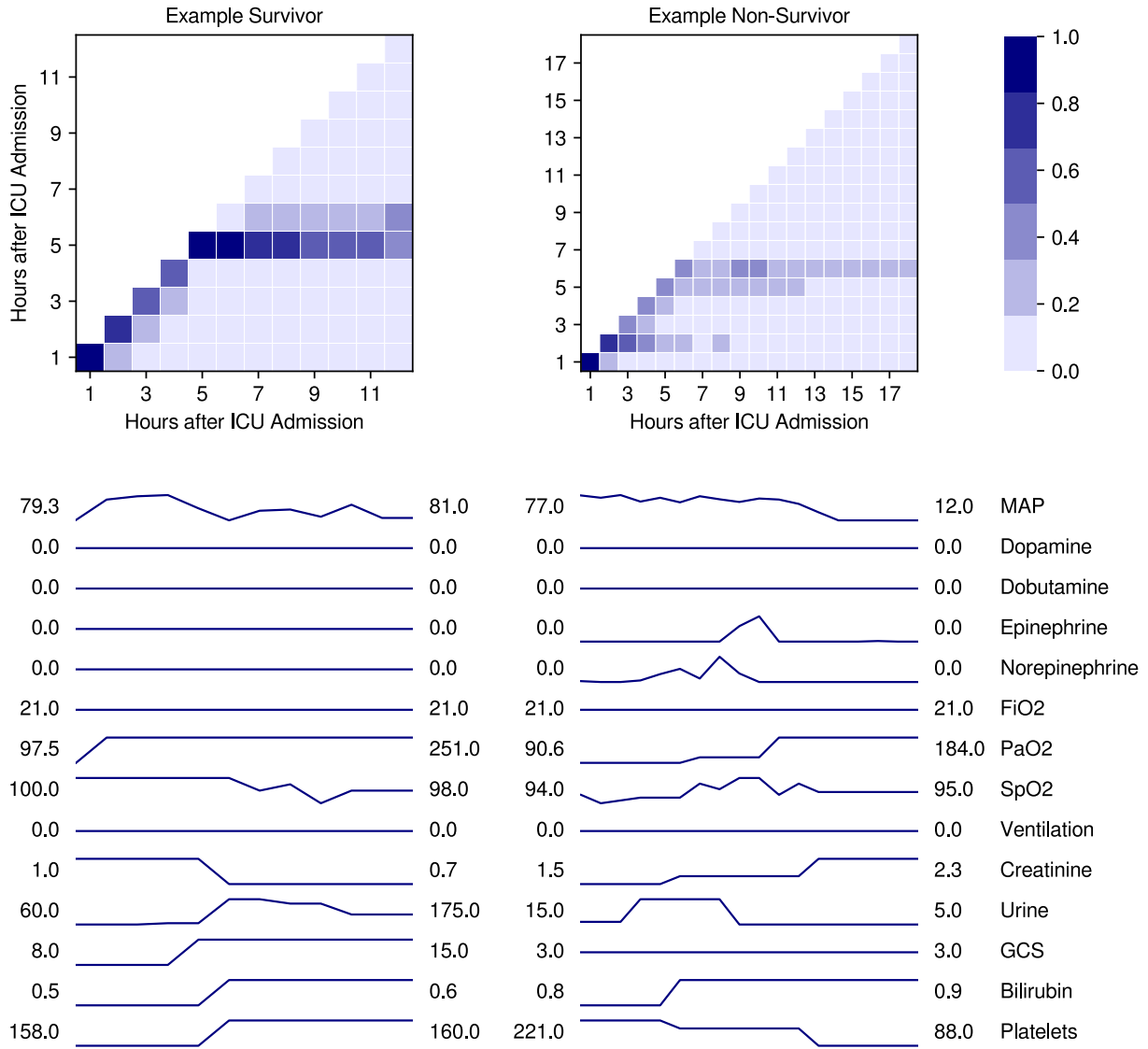

**Figure 5.** Visualized self-attention distributions for an example survivor and non-survivor from the *UFHealth* cohort, using DeepSOFA trained on the *MIMIC* cohort. Darker squares indicate increased model focus as a function of the passage of time (x-axis). Shown also are variable time series at each hour of the ICU stay, with initial and final measurement values shown on the left and right, respectively. MAP: mean arterial pressure, FiO2: fraction of inspired oxygen, PaO2: partial pressure of oxygen, SpO2: oxygen saturation, GCS: Glasgow Coma Scale.

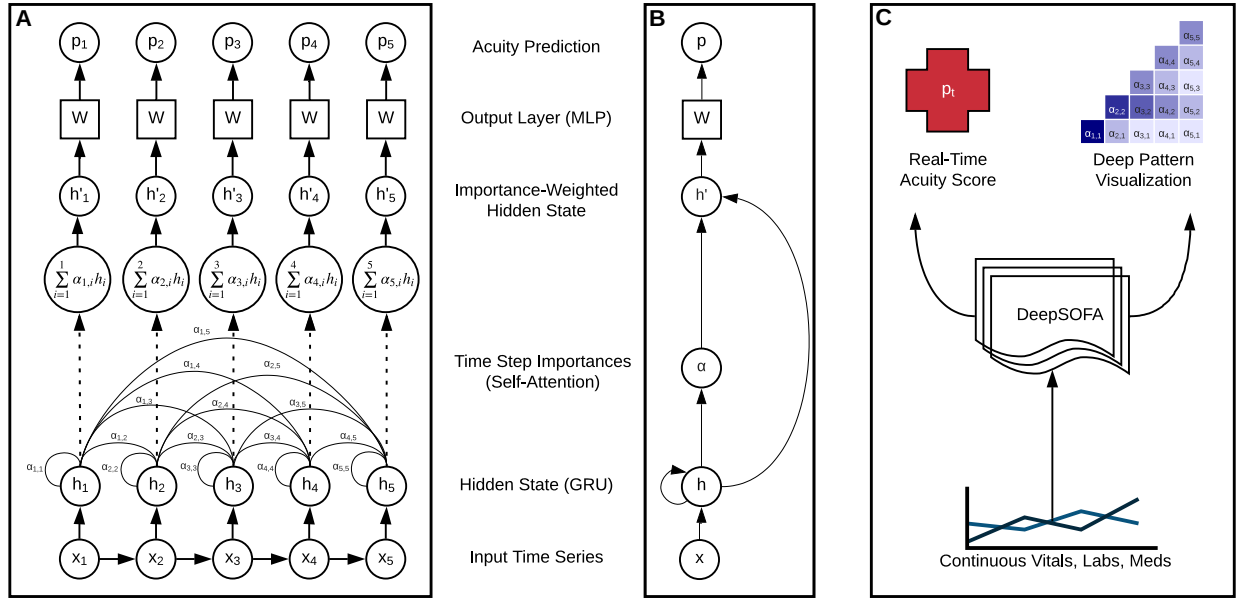

**Figure 6.** Three identical views of our DeepSOFA model at increasing levels of abstraction, from most technical (A) to highest level operation (C). Panel A illustrates mortality prediction calculation for an example data segment *x* of five hours in the ICU and corresponding hourly acuity assessments *p*, where at each time point only the current and previous time steps are used in attention calculations and mortality predictions. Panel B displays a more general and compact form for the same stages of data transformation. Panel C describes the high-level inputs and outputs for DeepSOFA, where along with overall acuity assessment, interpretable prediction rationale by way of salient sequence patterns are visualized by a heatmap of self-attention weights. A more technical description of each stage of DeepSOFA can be found in the online supplemental section *Model Details*.

# Supplementary Material

**Model Details**

**Recurrent neural network (RNN)**
For making hourly acuity assessments and mortality predictions, DeepSOFA utilizes a recurrent neural network (RNN), a type of deep learning algorithm that is naturally suited for processing sequential data. The key attribute of an RNN which makes it especially useful for modeling temporal data is its notion of internal memory; as each new time step of a sequence is processed, an RNN updates its hidden state by combining the current step's data with the deep representation of all data it has seen in the past. In its simplest form, the calculation of the RNN internal state is shown in Equation S1, where $U \in \mathbb{R}^{k \times d}$ is the input weight matrix, $V \in \mathbb{R}^{k \times k}$ is the recurrent weight matrix, $d$ is the number of features in the input sequence, $k$ is the tunable dimensionality of the RNN's hidden state, and bias terms are omitted for simplicity.

$$h_t = \sigma(Ux_t + Vh_{t-1}) \quad (S1)$$

At each time step $t$, the corresponding slice of the input sequence $x_t \in \mathbb{R}^d$ is combined with the previous time step's hidden state $h_{t-1} \in \mathbb{R}^k$ via $U, V$, and a nonlinear activation function $\sigma$ such as tanh. In practice, $h_0$ is often initialized to a vector of zeros. Similar to most other current deep learning algorithms, an RNN's weights are trained via backpropagation, a technique for updating a model's internal weights by attributing the final model error to individual parameters based on the flow of gradients from a differentiable loss function.

At any point during processing a sequence, the RNN's current hidden state $h_t$ is the representation of the entire sequence up to the current time. Once the sequence has been fully passed through the RNN, the final hidden state is taken as the deep representation of the entire sequence, which can then be passed to subsequent layers for tasks such as classification. Figure S1 shows two perspectives of the same example recurrent neural network.

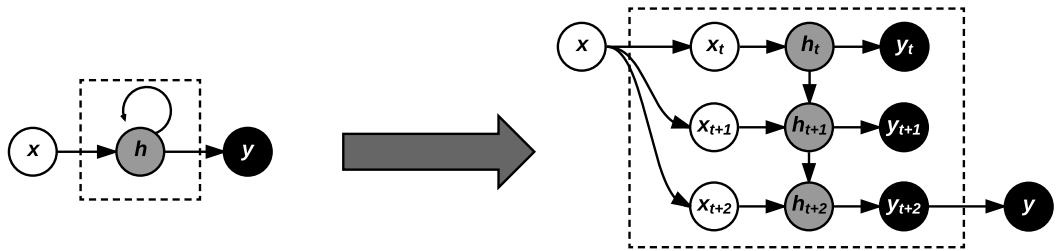

**Supplementary Figure S1.** Compact (left) and expanded (right) views of a recurrent neural network (RNN). For simplicity, we omit the fully-connected layer(s) typically placed between hidden state h$_t$ and prediction y$_t$. In this figure, y is taken to be the final prediction of interest for e.g. a traditional sequence classification task, but DeepSOFA uses all time step predictions y$_t$ for real-time acuity assessment and mortality prediction at every hour following ICU admission.

For predicting a classification target (such as DeepSOFA's in-hospital mortality) at a given time $t$, one would pass the RNN's current hidden state through a final classification layer as in Equation S2, where $W_y \in \mathbb{R}^{M \times k}$ is the output weight matrix, $M$ is the number of available prediction classes (two in DeepSOFA's mortality prediction), $k$ is the tunable dimensionality of the RNN's hidden state, $h_t$ is the RNN's internal hidden state at time $t$, and the bias term is omitted for simplicity.

$$y_t = W_y h_t \quad (S2)$$

**Gated recurrent units (GRU)**
In the previous section, we described the simplest form of a recurrent neural network. In practice, these standard RNNs suffer from what is known as the *exploding gradient problem*, in which repeated multiplications and nonlinear activations involved in updating the hidden state over time result in unstable chain rule-based gradient calculations during the backpropagation process. This issue, combined with the simple form of an RNN's internal memory, often results in non-robust models that are incapable of dealing with longer sequences and at best fail to utilize patterns or dependencies from distant past time steps.

In practice, two popular RNN modifications are typically preferred to the standard RNN, and both attempt to solve the issues above using slightly different techniques. The first RNN variant to be widely adopted is known as *long short-term memory* (LSTM), which augments the traditional RNN with additional weight matrices and gating functions to improve long-range sequential processing. More recently, *gated recurrent units* (GRU) have gained in popularity as a comparable alternative to the LSTM, again involving the introduction of new weights and operations for internally processing a sequence. While research has shown that these two methods result in similar performance for most tasks, DeepSOFA uses GRU networks due to the fewer number of required parameters. Equations S3-S6[1] illustrate the modifications to the traditional RNN from Equation S1, including the introduction of a "reset gate" $r_t$, an "update gate" $z_t$, and use of the elementwise multiplication operation $\odot$. Bias terms are omitted for simplicity.

$$
\begin{aligned}
r_t &= \sigma(W_r x_t + U_r h_{t-1}) &\quad \text{(S3)} \\
z_t &= \sigma(W_z x_t + U_z h_{t-1}) &\quad \text{(S4)} \\
h'_t &= \phi(W_x x_t + r_t \odot (U h_{t-1})) &\quad \text{(S5)} \\
h_t &= (1 - z_t) h_{t-1} + z_t h'_t &\quad \text{(S6)}
\end{aligned}
$$

In essence, the introduction of gating mechanisms allows the RNN to become more discretionary in the information it learns and remembers, and the modifications expand the capacity of its internal memory to incorporate important data from potentially long-distant time steps.

**Self-attention**

At each hour during an ICU stay, DeepSOFA makes a mortality probability calculation based on the sequence of EHR measurements available through the current hour. The simplest option for a given hour $t$ would involve passing the GRU's current hidden state $h_t$ through a final output classification layer to produce $y_t$ (Equation S2), the probability of in-hospital mortality given the sequence of measurements encountered thus far. This approach for temporal classification assumes that at a given time $t$, the entirety of information contained in the input sequence up through time $t$ can be fully represented by the fixed-dimensional vector $h_t$.

Rather than relying on the most recent hidden state of the GRU for making a prediction, we instead provide a weighted average of all prior hidden states to the final classification layer. The advantages to this approach are twofold. First, the sequence in effect becomes more distilled, by dampening inconsequential time steps and amplifying important ones relative to the final outcome. Second, and most importantly to DeepSOFA, these scalar time step weights can be used for providing clinician insight into the internal reasoning of the GRU's predictions, since larger timestep weights directly influences the averaged hidden state used for prediction and can thus be interpreted as denoting important time steps relative to the outcome of interest.

In the deep learning community, the process of learning scalar time step values to weight a sequence for saliency falls under the umbrella category of an *attention mechanism*, named as such in reference to the act of a model "focusing" on particular pieces of an input rather than its entirety. As is typical, in DeepSOFA we impose the constraint that all attention weights for a given sequence sum to 1 via use of a softmax function over the time dimension, thus encouraging larger weights being placed on the most important time steps of input data. Equations S7 and S8 illustrate a basic attention mechanism involving a global attention vector $W_{att} \in \mathbb{R}^{1 \times k}$, where relative importance (known in literature as compatibility) of each timestep's hidden state $h_i \in \mathbb{R}^{k \times 1}$ is calculated via dot product with the global attention vector, and a softmax function over compatibility scores is applied to yield a scalar weight $\alpha_i$ for each timestep $i = 1, 2, \ldots, t$, where the new sequence representation at a given time $t$ is computed as a weighted sum of all preceding hidden states.

$$
\begin{aligned}
\alpha_i &= \text{softmax}(W_{att} h_i) &\quad \text{(S7)} \\
h_t &= \sum_{i=0}^{t} \alpha_i h_i &\quad \text{(S8)}
\end{aligned}
$$

---

[1] https://arxiv.org/pdf/1412.3555.pdf

Attention mechanisms have garnered increased popularity in recent years, most notably in the field of natural language processing for tasks such as machine translation. In DeepSOFA, we adapt the more recent notion of *self-attention*[2], where rather than learning a global attention vector as in Equation S7, we directly calculate time step compatibilities between hidden time steps themselves. Since DeepSOFA is centered around real-time ICU monitoring, at each hour $t$ during a patient's ICU stay we wish to understand which prior timesteps were most influential in generating the current representation $h_t$ and current prediction $y_t$. By adopting a variant of self-attention, clinicians are able to understand the interactions between changes in measurements in real-time. Equations S9 and S10 describe DeepSOFA's attention process, where weight matrices $W_Q \in \mathbb{R}^{k \times k}$, $W_K \in \mathbb{R}^{k \times k}$, and $W_V \in \mathbb{R}^{k \times k}$ are learned projections of the same hidden representation $h_i \in \mathbb{R}^{k \times 1}$.

$$\alpha_i = \text{softmax}(W_Q h_i \cdot W_K h_i) \quad \text{(S9)}$$
$$h_t = \sum_{i=0}^{t} \alpha_i W_V h_i \quad \text{(S10)}$$

At each timestep $t$ during a patient's ICU stay, attention values $\alpha_i \ \forall \ i = 1,..t$ are recalculated to present a current, updated view on the most important time steps influencing mortality prediction. Where traditionally temporal attention values are calculated only after the entire sequence has been seen, to our knowledge DeepSOFA is the first framework to consider real-time self-attention distributions that are updated on-the-fly and only consider currently available EHR information for immediate clinician interpretability.

**Model training**
The architecture of the DeepSOFA model is relatively straightforward, consisting of (1) an input layer, (2) a single GRU layer, (3) a self-attention layer, (4) a dropout layer, and (5) a final fully-connected output layer for classification. Since DeepSOFA is focused on real-time prediction, we replicate the final mortality target[3] across all input time steps to encourage a correct prediction as early as possible. The model is trained to optimize cross-entropy loss averaged across all hours of an ICU stay (Equation S11), where $W_Y$ is the output weight matrix of the final classification layer, $h_t$ is the attention-weighted sum of available hidden states at each hour $t \ \forall \ t = 1,...T$, and $T$ is the number of hours the patient was in the ICU. Since we apply target replication, $y$ is the same for all time steps $t$ for a given ICU stay.

$$loss = \frac{1}{T}\sum_{t=0}^{T} -[y * \log(W_Y h_t) + (1-y)\log(1 - W_Y h_t)] \quad \text{(S11)}$$

Model training was terminated when the AUC on the validation set did not increase for 5 epochs. DeepSOFA was specifically trained with a focus on real-time implementation; thus, during training we implemented the dynamic, hourly self-attention updates by copying the input sequence of length $T$ to a $T \times T$ matrix and applying a lower triangular mask for preventing the model from looking ahead to time steps occurring in the future.

**Model parameters**
In this iteration of our work, we did not perform extensive hyperparameter tuning and simply sought to show that recurrent neural networks, combined with the full continuous scope of a subset of EHR data, outperformed clinical baselines while providing more granular and informative time series patterns to improve ICU bedside monitoring. This initial version of DeepSOFA used a GRU layer with 64 hidden units, 20% dropout, L2 weight regularization of 1e-6, a batch size of 16, and an Adam optimizer.

---

[2] http://papers.nips.cc/paper/7181-attention-is-all-you-need.pdf
[3] https://arxiv.org/pdf/1511.03677.pdf

**Supplementary tables and figures**

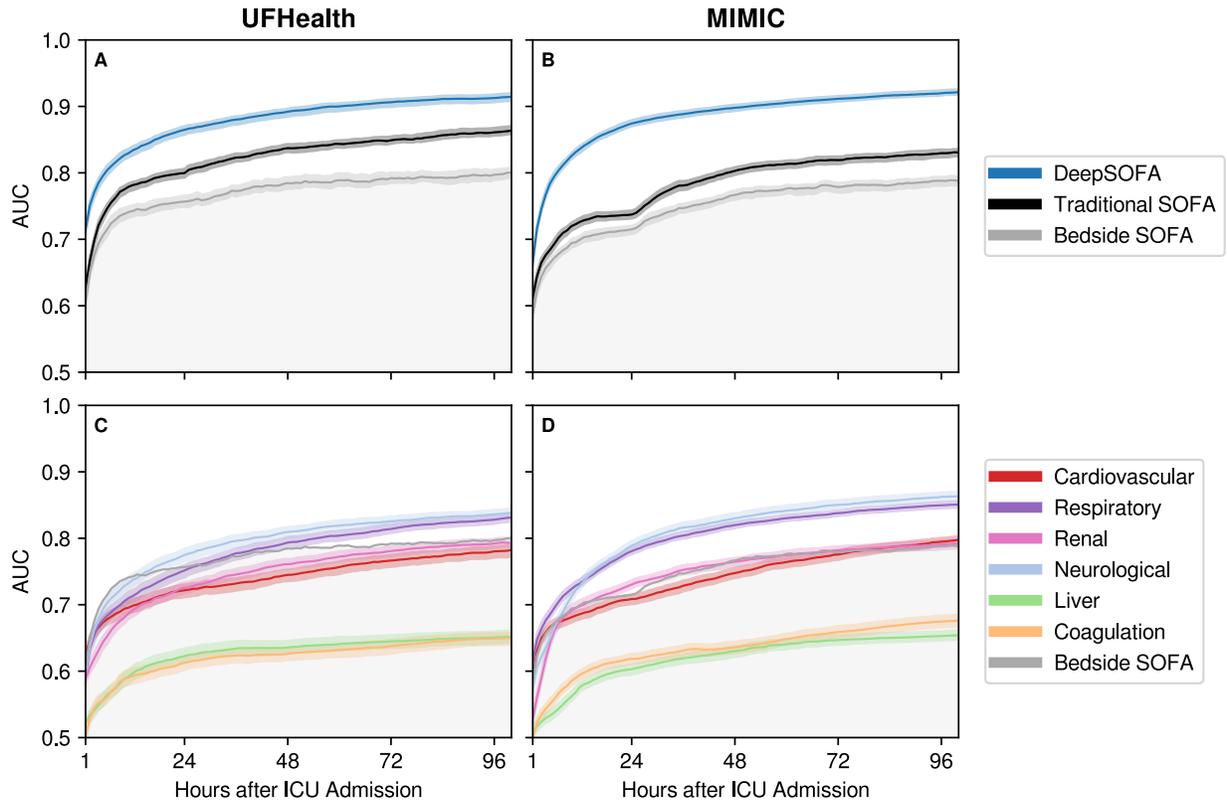

**Supplementary Figure S2.** (A, B) Internally validated DeepSOFA, Bedside SOFA, and Traditional SOFA score accuracy in predicting in-hospital mortality, expressed as area under the receiver operating characteristic curve (AUC) for the first 100 hours following ICU admission. (C, D) Internally validated DeepSOFA accuracy for individual models corresponding to variable sets derived from SOFA organ system classification for the first 100 hours following ICU admission. Shaded regions represent 95% confidence intervals based on 100 bootstrapped iterations. All internal validation results obtained using 5-fold cross-validation. Columns specify both the development and validation cohort. SOFA: Sequential Organ Failure Assessment.

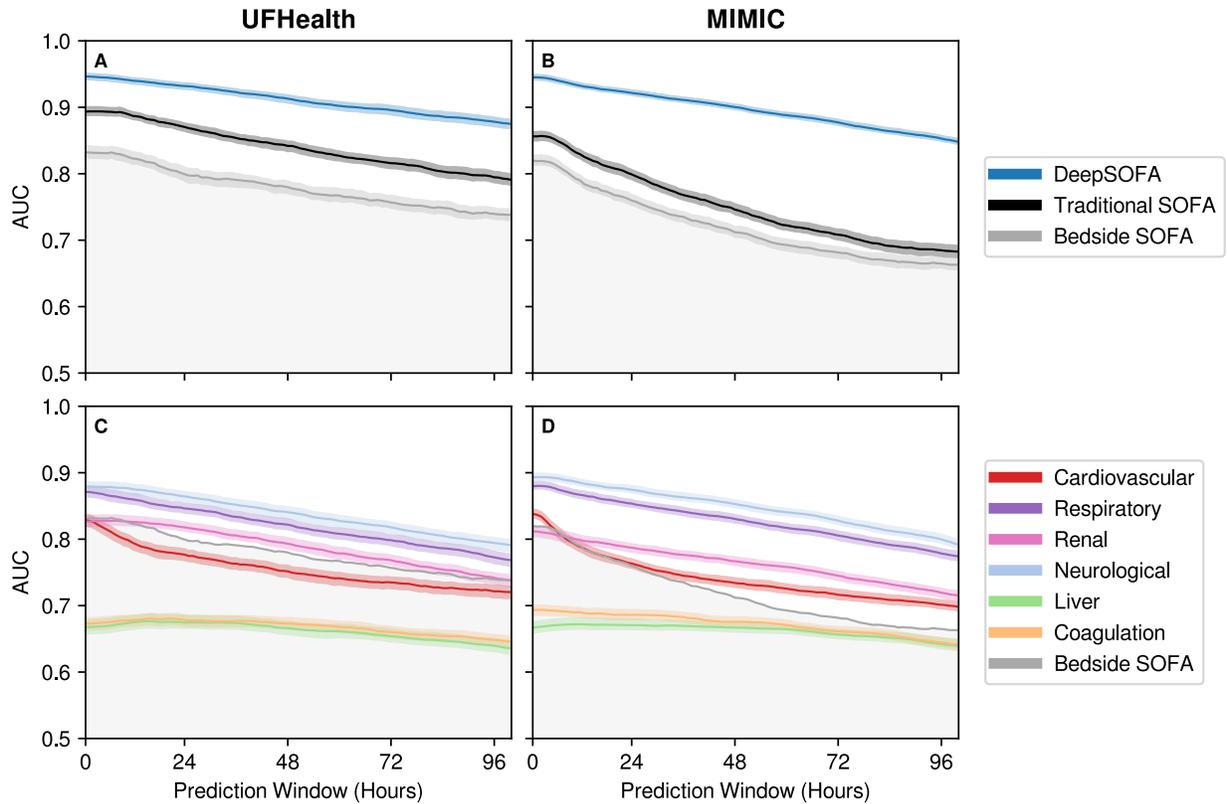

**Supplementary Figure S3.** (A, B) Internally validated DeepSOFA, Bedside SOFA, and Traditional SOFA score accuracy in predicting in-hospital mortality, expressed as area under the receiver operating characteristic curve (AUC) for 100 hours preceding death or hospital discharge. (C, D) Internally validated DeepSOFA accuracy for individual models corresponding to variable sets derived from SOFA organ system classification for the 100 hours preceding death or hospital discharge. Shaded regions represent 95% confidence intervals based on 100 bootstrapped iterations. All internal validation results obtained using 5-fold cross-validation. Columns specify both the development and validation cohort. SOFA: Sequential Organ Failure Assessment.

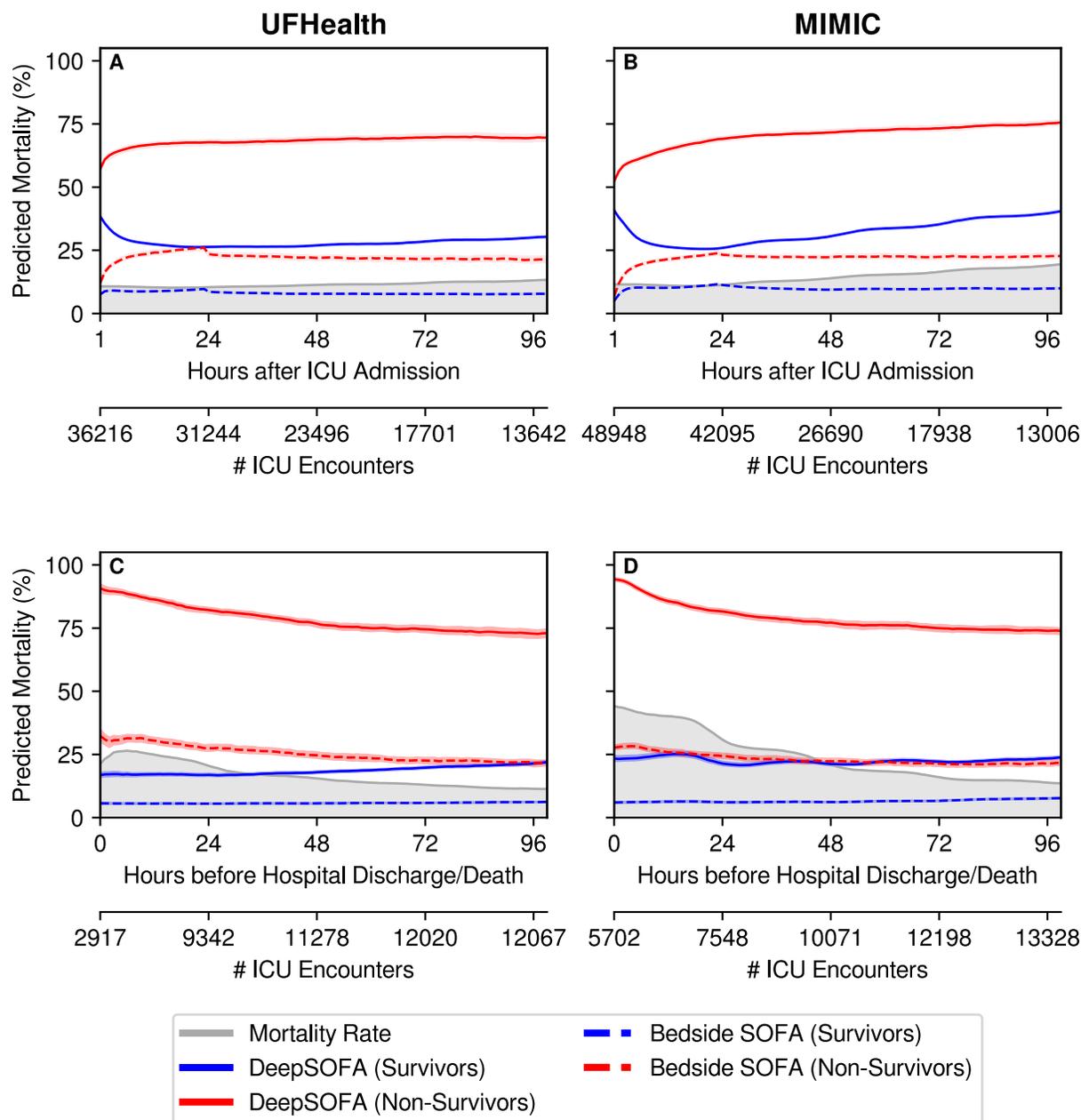

**Supplementary Figure S4.** Mean predicted mortality probabilities for internally validated DeepSOFA and Bedside SOFA models stratified by outcome. Probabilities shown both for first 100 hours after ICU admission (A, B) and final 100 hours before hospital discharge or death (C, D). Number of ongoing ICU encounters shown below each panel. Shaded regions around each line represent 95% confidence intervals based on 100 bootstrapped iterations. Gray shared area denotes hourly mortality rate for active ICU encounters. All internal validation results obtained using 5-fold cross-validation. Columns specify both the development and validation cohort. SOFA: Sequential Organ Failure Assessment.

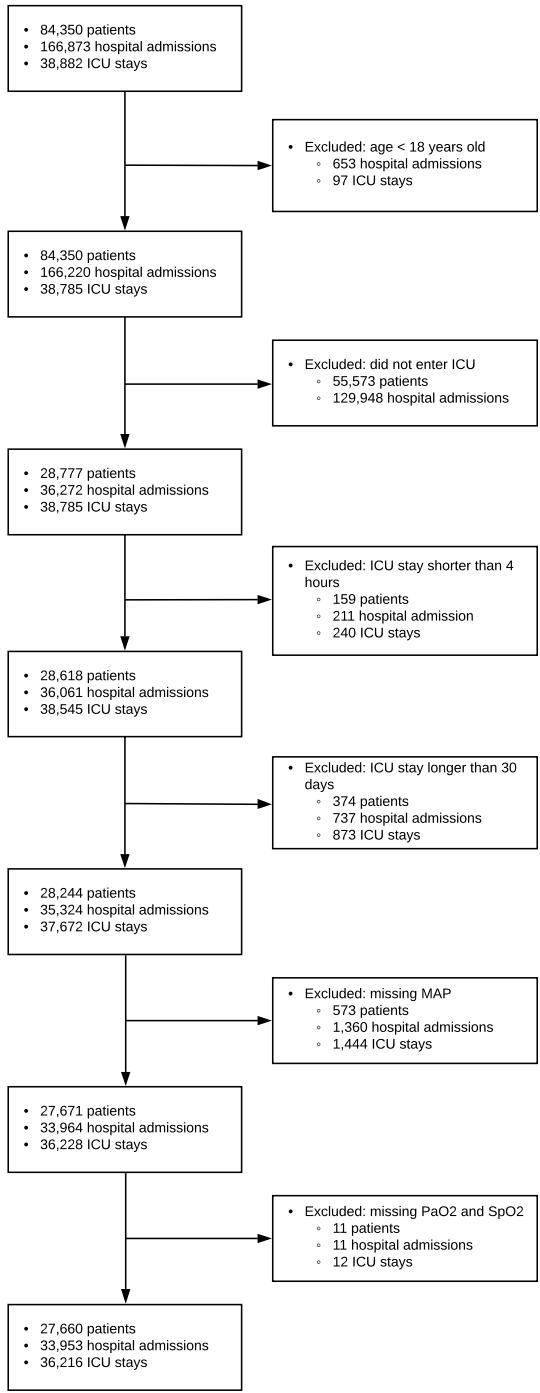
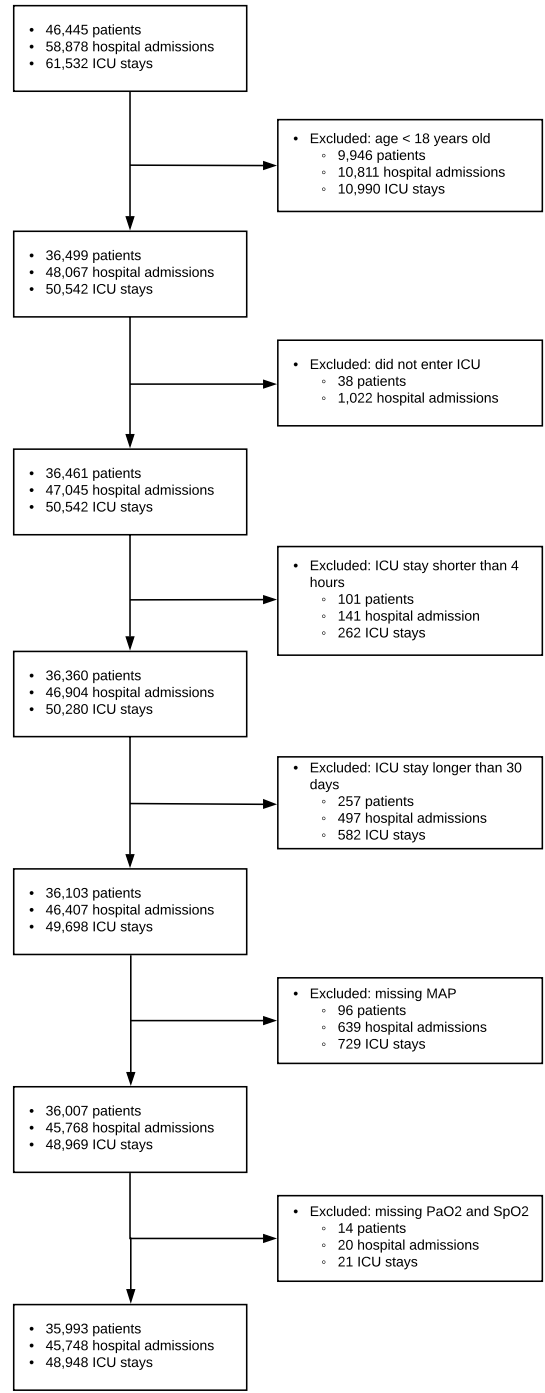

**Supplementary Figure S5.** Cohort selection and exclusion criteria for both *UFHealth* and *MIMIC* cohorts.

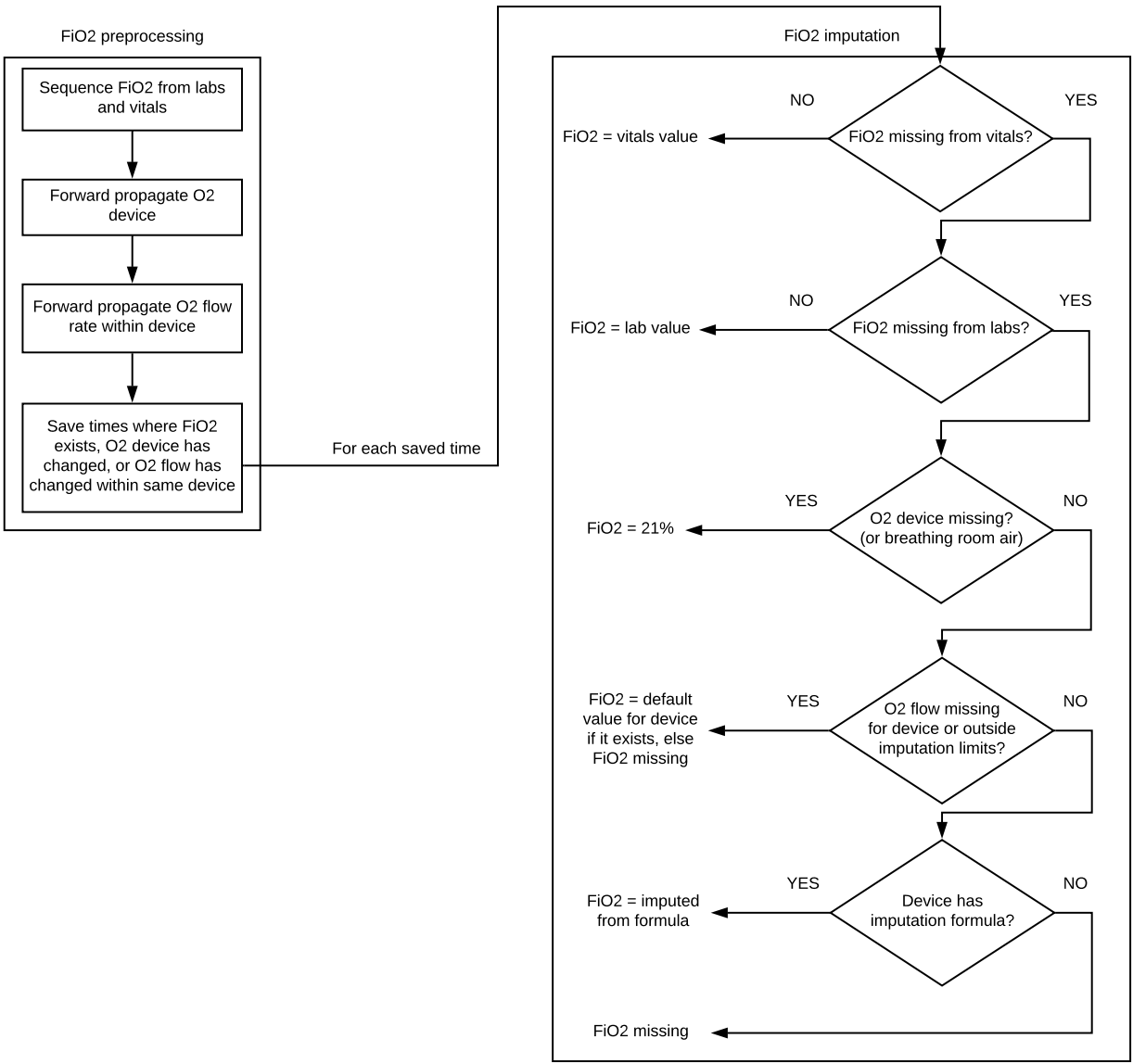

**Supplementary Figure S6.** Process for FiO2 imputation from existing vitals or laboratory measurements, oxygen delivery device, and corresponding oxygen flow rate.

| Device | Default FiO2 | O2 limits for imputation [min, max] L/min | FiO2 imputation formula X: O2 flow, L/min | Max imputed FiO2 |
|---|---|---|---|---|
| Aerosol mask | 35 | [0, N/A] | 21 + (X * 4) | 60 |
| Nasal cannula | ----- | [0, N/A] | 21 + (X * 4) | 40 |
| High flow nasal cannula | 50 | [6, 15] | 48 + [(X - 6) * 2] | 100 |
| Simple mask | ----- | [0, 19] | 21 + (X * 4) | 60 |
| Non-rebreather mask | 60 | [8, N/A] | 80 + [minimum(X - 10, 2) * 10] | 100 |
| Venturi mask | 35 | [4, 8] | 26 + [(X - 4) * 2.5] | 55 |
| Trach mask | 30 | ----- | ----- | ----- |
| CPAP | 40 | ----- | ----- | ----- |
| BiPAP | 40 | ----- | ----- | ----- |
| Tracheostomy | 40 | ----- | ----- | ----- |
| Ventilator | 40 | ----- | ----- | ----- |
| Bag valve mask | 100 | ----- | ----- | ----- |
| T-piece | 40 | ----- | ----- | ----- |
| Transtracheal catheter | 40 | ----- | ----- | ----- |
| Blow-by | 25 | ----- | ----- | ----- |
| Partial rebreather mask | 35 | ----- | ----- | ----- |
| Face tent | 25 | ----- | ----- | ----- |
| Oxyimiser | 40 | ----- | ----- | ----- |
| Oscillator | 80 | ----- | ----- | ----- |
| Oxyhood | 35 | ----- | ----- | ----- |

**Supplementary Table S1.** Formulas for imputing FiO2 from oxygen delivery device and corresponding oxygen flow rate. If no oxygen flow rate is given, default FiO2 is imputed. If oxygen flow rate is outside specified range, minimum and maximum flow rate is used for imputing FiO2. If formula result is greater than maximum per-device FiO2, the maximum FiO2 is imputed. Abbreviations: CPAP, continuous positive airway pressure; BiPAP, bi-level positive airway pressure.

| SOFA component (organ system) | Tradtional SOFA | DeepSOFA | Variables, units of measurement | Non-outlier range [min, max] | Missing values (traditional SOFA)† |
|---|---|---|---|---|---|
| Cardiovascular | 4: Dop > 15 or epi > 0.1 or nor > 0.1<br>3: Dop > 5 or epi <= 0.1<br>2: Dop <= 5 or dob (any dose)<br>1: MAP < 70<br>0: MAP >= 70 and no vasopressors | Uses all available MAP recordings and drug dosing for dop, dob, epi, and nor | Mean arterial blood pressure, mmHg | (0, 300] | SOFA component = 0 |
| | | | Dopamine, mcg/kg/min | [0, 50] | N/A (normal SOFA calculation) |
| | | | Dobutamine, mcg/kg/min | [0, 40] | N/A (normal SOFA calculation) |
| | | | Epinephrine, mcg/kg/min | [0, 5] | N/A (normal SOFA calculation) |
| | | | Norepinephrine, mcg/kg/min | [0, 15] | N/A (normal SOFA calculation) |
| Respiratory | 4: PaO2/FiO2 < 100 w/ MV<br>3: PaO2/FiO2 < 200 w/ MV<br>2: PaO2/FiO2 < 300<br>1: PaO2/FiO2 < 400<br>0: PaO2/FiO2 >= 400 | Uses all available PaO2, sPO2, FiO2, and mechanical ventilation status recordings | Fraction of inspired oxygen (FiO2), % | [21, 100] | SOFA component = 0 |
| | | | Partial pressure of oxygen (PaO2), mmHg | (0, 800] | Use SpO2/FiO2 to PaO2/FiO2 conversion |
| | | | Blood oxygen saturation (SpO2), % | (0, 100] | If PaO2 also missing, SOFA component = 0. Otherwise, normal SOFA calculation |
| | | | Mechanical ventilation, binary indicator | N/A | N/A (normal SOFA calculation) |
| Central nervous system | 4: GCS < 6<br>3: GCS < 10<br>2: GCS < 13<br>1: GCS < 15<br>0: GCS = 15 | Uses all available GCS recordings | Glasgow coma scale, score | [3,15] | SOFA component = 0 |
| Coagulation | 4: Platelets < 20<br>3: Platelets < 50<br>2: Platelets < 100<br>1: Platelets < 150<br>0: Platelets >= 150 | Uses all available platelet count recordings | Platelets, count x 10^3/mm^3 | (0, 832]* | SOFA component = 0 |
| Liver | 4: Bilirubin > 12<br>3: Bilirubin >= 6<br>2: Bilirubin >= 2<br>1: Bilirubin >= 1.2<br>0: Bilirubin < 1.2 | Uses all available bilirubin recordings | Bilirubin, mg/dL | (0, 50] | SOFA component = 0 |
| Renal | 4: Creatinine > 5 or urine sum < 200<br>3: Creatinine >= 3.5 or urine sum < 500<br>2: Creatinine <= 2<br>1: Creatinine <= 1.2<br>0: Creatinine < 1.2 | Uses all available creatinine and urine recordings | Creatinine, mg/dL | (0, 30] | SOFA component = 0 |
| | | | Urine, mL | [0, 1095]* | Normal SOFA calculation |

**Supplementary Table S2.** SOFA score, DeepSOFA, variable, and processing definitions. *Threshold values obtained from modified Z-score method (Z > 5). †DeepSOFA model works with all available data, even if some variables are missing. Abbreviations: Dop, dopamine; Dob, dobutamine; Epi, epinephrine; nor, norepinephrine; MAP, mean arterial pressure; MV, mechanical ventilation.

| Model | Training Required? | Mortality Probability Prediction at Hour T | AUC Calculation at Hour T |
|---|---|---|---|
| DeepSOFA | Yes | Feed current data sequence from hour 0 to hour T through RNN to get mortality probability prediction. | Get mortality probability predictions at hour T for all ICU encounters in validation cohort. For ICU encounters already completed by hour T, use final mortality prediction. Calculate AUC between mortality probability predictions and true in-hospital mortality labels. |
| Bedside SOFA | No | Calculate SOFA score for previous 24-hour period (T-24, T]. Consult published mortality rate table relating SOFA score to overall mortality rate. Use mortality rate as patient mortality probability prediction. | |
| Traditional SOFA | | N/A | Calculate SOFA score for previous 24-hour period (T-24, T] for all ICU encounters in validation cohort. For ICU encounters already completed by hour T, use final SOFA score. Calculate AUC between SOFA scores and true in-hospital mortality labels. |

**Supplementary Table S3.** High-level overview of DeepSOFA and SOFA baseline model operation.

| | | First | 24h | 48h | 72h | 96h | Last | Mean |
|---|---|---|---|---|---|---|---|---|
| **UFHealth** | DeepSOFA | 0.86 (0.86 - 0.87) | 0.89 (0.88 - 0.90) | 0.91 (0.90 - 0.91) | 0.91 (0.91 - 0.92) | 0.72 (0.71 - 0.72) | 0.95 (0.94 - 0.95) | 0.91 (0.90 - 0.91) |
| | Logistic Regression | 0.77 (0.77 - 0.78) | 0.80 (0.80 - 0.81) | 0.82 (0.81 - 0.83) | 0.83 (0.82 - 0.84) | 0.61 (0.60 - 0.62) | 0.88 (0.87 - 0.89) | 0.81 (0.81 - 0.82) |
| | Random Forest | 0.68 (0.67 - 0.69) | 0.71 (0.70 - 0.71) | 0.72 (0.72 - 0.73) | 0.74 (0.73 - 0.75) | 0.57 (0.57 - 0.58) | 0.81 (0.81 - 0.82) | 0.72 (0.71 - 0.73) |
| **MIMIC** | DeepSOFA | 0.87 (0.87 - 0.88) | 0.90 (0.89 - 0.90) | 0.91 (0.91 - 0.92) | 0.92 (0.92 - 0.92) | 0.67 (0.66 - 0.67) | 0.95 (0.94 - 0.95) | 0.91 (0.91 - 0.92) |
| | Logistic Regression | 0.77 (0.76 - 0.77) | 0.80 (0.80 - 0.81) | 0.82 (0.82 - 0.83) | 0.83 (0.83 - 0.84) | 0.57 (0.56 - 0.57) | 0.87 (0.87 - 0.88) | 0.81 (0.80 - 0.81) |
| | Random Forest | 0.67 (0.67 - 0.68) | 0.71 (0.70 - 0.71) | 0.73 (0.72 - 0.73) | 0.74 (0.74 - 0.75) | 0.54 (0.54 - 0.55) | 0.81 (0.81 - 0.82) | 0.72 (0.72 - 0.73) |

**Supplementary Table S4.** Internally validated AUC results for three additional baseline machine learning models compared with DeepSOFA results from main manuscript. Baseline models used 84 aggregate features recalculated at every hour after ICU admission, including the following for each of the 14 SOFA variables: minimum value, maximum value, mean value, standard deviation, first value, and last value. Models trained using aggregate features from entire development ICU stays and evaluated hourly using recalculated features in an expanding window. Results shown for first hour after ICU admission, at time of ICU discharge, mean across all hours, and for 24, 48, 72, and 96 hours after ICU admission. 95% confidence intervals shown in parentheses.

| | | First | 24h | 48h | 72h | 96h | Last | Mean |
|---|---|---|---|---|---|---|---|---|
| **UFHealth** | DeepSOFA | 0.85 (0.85 - 0.86) | 0.89 (0.88 - 0.89) | 0.90 (0.89 - 0.91) | 0.91 (0.90 - 0.92) | 0.70 (0.70 - 0.71) | 0.94 (0.94 - 0.95) | 0.90 (0.90 - 0.91) |
| | Logistic Regression | 0.76 (0.76 - 0.77) | 0.79 (0.79 - 0.80) | 0.81 (0.81 - 0.82) | 0.83 (0.82 - 0.83) | 0.60 (0.59 - 0.60) | 0.87 (0.87 - 0.88) | 0.80 (0.80 - 0.81) |
| | Random Forest | 0.63 (0.62 - 0.63) | 0.65 (0.64 - 0.66) | 0.68 (0.67 - 0.68) | 0.69 (0.69 - 0.70) | 0.53 (0.52 - 0.53) | 0.77 (0.77 - 0.78) | 0.67 (0.67 - 0.68) |
| **MIMIC** | DeepSOFA | 0.86 (0.85 - 0.86) | 0.89 (0.88 - 0.89) | 0.90 (0.90 - 0.90) | 0.91 (0.91 - 0.91) | 0.61 (0.61 - 0.62) | 0.94 (0.93 - 0.94) | 0.90 (0.90 - 0.90) |
| | Logistic Regression | 0.75 (0.74 - 0.75) | 0.79 (0.78 - 0.79) | 0.80 (0.80 - 0.81) | 0.82 (0.81 - 0.82) | 0.58 (0.57 - 0.58) | 0.86 (0.85 - 0.86) | 0.79 (0.79 - 0.80) |
| | Random Forest | 0.66 (0.66 - 0.67) | 0.69 (0.69 - 0.70) | 0.72 (0.71 - 0.72) | 0.73 (0.72 - 0.74) | 0.54 (0.54 - 0.55) | 0.81 (0.81 - 0.82) | 0.71 (0.70 - 0.72) |

**Supplementary Table S5.** Externally validated AUC results for three additional baseline machine learning models compared with DeepSOFA results from main manuscript. Baseline models used 84 aggregate features recalculated at every hour after ICU admission, including the following for each of the 14 SOFA variables: minimum value, maximum value, mean value, standard deviation, first value, and last value. Models trained using aggregate features from entire development ICU stays and evaluated hourly using recalculated features in an expanding window. Results shown for first hour after ICU admission, at time of ICU discharge, mean across all hours, and for 24, 48, 72, and 96 hours after ICU admission. 95% confidence intervals shown in parentheses.

| | | | First | 24h | 48h | 72h | 96h | Last | Mean |
|---|---|---|---|---|---|---|---|---|---|
| UFHealth | DeepSOFA | All | 0.86 (0.86 - 0.87) | 0.89 (0.88 - 0.90) | 0.91 (0.90 - 0.91) | 0.91 (0.91 - 0.92) | 0.72 (0.71 - 0.72) | 0.95 (0.94 - 0.95) | 0.91 (0.90 - 0.91) |
| | | First Only | 0.87 (0.87 - 0.88) | 0.90 (0.89 - 0.91) | 0.92 (0.91 - 0.92) | 0.92 (0.92 - 0.93) | 0.72 (0.71 - 0.73) | 0.95 (0.95 - 0.96) | 0.91 (0.90 - 0.91) |
| | | Unique Only | 0.89 (0.89 - 0.90) | 0.92 (0.92 - 0.93) | 0.94 (0.93 - 0.94) | 0.94 (0.94 - 0.95) | 0.73 (0.72 - 0.74) | 0.97 (0.97 - 0.97) | 0.93 (0.92 - 0.93) |
| | Traditional SOFA | All | 0.80 (0.79 - 0.81) | 0.84 (0.83 - 0.84) | 0.85 (0.84 - 0.85) | 0.86 (0.86 - 0.87) | 0.63 (0.62 - 0.63) | 0.89 (0.89 - 0.90) | 0.85 (0.85 - 0.86) |
| | | First Only | 0.81 (0.80 - 0.82) | 0.85 (0.84 - 0.86) | 0.86 (0.85 - 0.87) | 0.87 (0.87 - 0.88) | 0.63 (0.62 - 0.64) | 0.91 (0.90 - 0.91) | 0.85 (0.85 - 0.86) |
| | | Unique Only | 0.83 (0.82 - 0.83) | 0.87 (0.86 - 0.88) | 0.88 (0.88 - 0.89) | 0.89 (0.89 - 0.90) | 0.63 (0.62 - 0.64) | 0.93 (0.92 - 0.93) | 0.87 (0.87 - 0.88) |
| | Bedside SOFA | All | 0.76 (0.75 - 0.77) | 0.78 (0.78 - 0.79) | 0.79 (0.78 - 0.80) | 0.80 (0.79 - 0.80) | 0.61 (0.60 - 0.62) | 0.83 (0.82 - 0.84) | 0.79 (0.79 - 0.80) |
| | | First Only | 0.77 (0.76 - 0.78) | 0.79 (0.79 - 0.80) | 0.81 (0.79 - 0.82) | 0.81 (0.80 - 0.82) | 0.61 (0.60 - 0.62) | 0.84 (0.83 - 0.85) | 0.80 (0.79 - 0.81) |
| | | Unique Only | 0.78 (0.77 - 0.79) | 0.81 (0.80 - 0.82) | 0.82 (0.82 - 0.83) | 0.83 (0.82 - 0.84) | 0.61 (0.60 - 0.62) | 0.86 (0.85 - 0.87) | 0.82 (0.81 - 0.83) |
| MIMIC | DeepSOFA | All | 0.87 (0.87 - 0.88) | 0.90 (0.89 - 0.90) | 0.91 (0.91 - 0.92) | 0.92 (0.92 - 0.92) | 0.67 (0.66 - 0.67) | 0.95 (0.94 - 0.95) | 0.91 (0.91 - 0.92) |
| | | First Only | 0.89 (0.88 - 0.89) | 0.91 (0.90 - 0.91) | 0.92 (0.91 - 0.92) | 0.93 (0.92 - 0.93) | 0.67 (0.66 - 0.68) | 0.96 (0.95 - 0.96) | 0.91 (0.91 - 0.92) |
| | | Unique Only | 0.91 (0.91 - 0.92) | 0.93 (0.93 - 0.94) | 0.95 (0.94 - 0.95) | 0.96 (0.95 - 0.96) | 0.68 (0.67 - 0.69) | 0.98 (0.97 - 0.98) | 0.94 (0.93 - 0.94) |
| | Traditional SOFA | All | 0.74 (0.73 - 0.74) | 0.80 (0.80 - 0.81) | 0.82 (0.81 - 0.82) | 0.83 (0.82 - 0.84) | 0.61 (0.61 - 0.62) | 0.86 (0.85 - 0.86) | 0.82 (0.81 - 0.82) |
| | | First Only | 0.74 (0.73 - 0.74) | 0.81 (0.80 - 0.81) | 0.83 (0.82 - 0.83) | 0.84 (0.83 - 0.84) | 0.61 (0.60 - 0.62) | 0.86 (0.86 - 0.87) | 0.81 (0.80 - 0.82) |
| | | Unique Only | 0.76 (0.75 - 0.76) | 0.83 (0.83 - 0.84) | 0.85 (0.84 - 0.86) | 0.86 (0.85 - 0.87) | 0.62 (0.61 - 0.63) | 0.89 (0.88 - 0.89) | 0.83 (0.83 - 0.84) |
| | Bedside SOFA | All | 0.71 (0.71 - 0.72) | 0.77 (0.76 - 0.77) | 0.78 (0.77 - 0.79) | 0.79 (0.78 - 0.79) | 0.59 (0.58 - 0.60) | 0.82 (0.81 - 0.83) | 0.78 (0.77 - 0.79) |
| | | First Only | 0.71 (0.70 - 0.72) | 0.77 (0.76 - 0.78) | 0.79 (0.78 - 0.79) | 0.80 (0.79 - 0.80) | 0.59 (0.58 - 0.60) | 0.83 (0.82 - 0.84) | 0.77 (0.77 - 0.78) |
| | | Unique Only | 0.73 (0.72 - 0.74) | 0.80 (0.79 - 0.80) | 0.81 (0.80 - 0.82) | 0.82 (0.81 - 0.83) | 0.59 (0.58 - 0.60) | 0.85 (0.85 - 0.86) | 0.80 (0.79 - 0.81) |

**Supplementary Table S6.** Internally validated AUC results for three variants of handling patients with multiple ICU stays. Results shown for first hour after ICU admission, at time of ICU discharge, mean across all hours, and for 24, 48, 72, and 96 hours after ICU admission. "All" denotes using all ICU stays available, and is the setting used in the main manuscript. For patients with multiple ICU stays in their EHR, "first only" denotes removing all but their first ICU stay available, while "unique only" removes these patients altogether, ignoring all of their ICU stays. For each setting, new models were trained and evaluated using the modified datasets. All internal validation results obtained using 5-fold cross-validation. 95% confidence intervals shown in parentheses. Modified ICU stay counts: (1) UFHealth: All = 36,216; First Only = 27,660; Unique Only = 22,229; (2) MIMIC: All = 45,748; First Only = 35,993; Unique Only = 28,445.

| | | | First | 24h | 48h | 72h | 96h | Last | Mean |
|---|---|---|---|---|---|---|---|---|---|
| UFHealth | DeepSOFA | All | 0.85 (0.85 - 0.86) | 0.89 (0.88 - 0.89) | 0.90 (0.89 - 0.91) | 0.91 (0.90 - 0.92) | 0.70 (0.70 - 0.71) | 0.94 (0.94 - 0.95) | 0.90 (0.90 - 0.91) |
| | | First Only | 0.87 (0.86 - 0.87) | 0.90 (0.89 - 0.90) | 0.91 (0.90 - 0.92) | 0.92 (0.91 - 0.92) | 0.71 (0.70 - 0.72) | 0.95 (0.94 - 0.95) | 0.90 (0.90 - 0.91) |
| | | Unique Only | 0.89 (0.88 - 0.90) | 0.92 (0.91 - 0.92) | 0.93 (0.93 - 0.94) | 0.94 (0.93 - 0.95) | 0.72 (0.71 - 0.73) | 0.97 (0.97 - 0.97) | 0.92 (0.92 - 0.93) |
| | Traditional SOFA | All | 0.80 (0.79 - 0.81) | 0.84 (0.83 - 0.84) | 0.85 (0.84 - 0.85) | 0.86 (0.86 - 0.87) | 0.63 (0.62 - 0.63) | 0.89 (0.89 - 0.90) | 0.85 (0.85 - 0.86) |
| | | First Only | 0.81 (0.80 - 0.82) | 0.85 (0.84 - 0.86) | 0.86 (0.85 - 0.87) | 0.87 (0.87 - 0.88) | 0.63 (0.62 - 0.64) | 0.91 (0.90 - 0.91) | 0.85 (0.85 - 0.86) |
| | | Unique Only | 0.83 (0.82 - 0.83) | 0.87 (0.86 - 0.88) | 0.88 (0.88 - 0.89) | 0.89 (0.89 - 0.90) | 0.63 (0.62 - 0.64) | 0.93 (0.92 - 0.93) | 0.87 (0.87 - 0.88) |
| | Bedside SOFA | All | 0.76 (0.75 - 0.77) | 0.78 (0.78 - 0.79) | 0.79 (0.78 - 0.80) | 0.80 (0.79 - 0.80) | 0.61 (0.60 - 0.62) | 0.83 (0.82 - 0.84) | 0.79 (0.79 - 0.80) |
| | | First Only | 0.77 (0.76 - 0.78) | 0.79 (0.79 - 0.80) | 0.81 (0.79 - 0.82) | 0.81 (0.80 - 0.82) | 0.61 (0.60 - 0.62) | 0.84 (0.83 - 0.85) | 0.80 (0.79 - 0.81) |
| | | Unique Only | 0.78 (0.77 - 0.79) | 0.81 (0.80 - 0.82) | 0.82 (0.82 - 0.83) | 0.83 (0.82 - 0.84) | 0.61 (0.60 - 0.62) | 0.86 (0.85 - 0.87) | 0.82 (0.81 - 0.83) |
| MIMIC | DeepSOFA | All | 0.86 (0.85 - 0.86) | 0.89 (0.88 - 0.89) | 0.90 (0.90 - 0.90) | 0.91 (0.91 - 0.91) | 0.61 (0.61 - 0.62) | 0.94 (0.93 - 0.94) | 0.90 (0.90 - 0.90) |
| | | First Only | 0.87 (0.86 - 0.87) | 0.90 (0.89 - 0.90) | 0.91 (0.91 - 0.92) | 0.92 (0.92 - 0.93) | 0.61 (0.60 - 0.62) | 0.95 (0.94 - 0.95) | 0.90 (0.89 - 0.90) |
| | | Unique Only | 0.90 (0.89 - 0.90) | 0.92 (0.92 - 0.93) | 0.94 (0.93 - 0.94) | 0.95 (0.94 - 0.95) | 0.62 (0.61 - 0.63) | 0.97 (0.97 - 0.98) | 0.93 (0.92 - 0.93) |
| | Traditional SOFA | All | 0.74 (0.73 - 0.74) | 0.80 (0.80 - 0.81) | 0.82 (0.81 - 0.82) | 0.83 (0.82 - 0.84) | 0.61 (0.61 - 0.62) | 0.86 (0.85 - 0.86) | 0.82 (0.81 - 0.82) |
| | | First Only | 0.74 (0.73 - 0.74) | 0.81 (0.80 - 0.81) | 0.83 (0.82 - 0.83) | 0.84 (0.83 - 0.84) | 0.61 (0.60 - 0.62) | 0.86 (0.86 - 0.87) | 0.81 (0.80 - 0.82) |
| | | Unique Only | 0.76 (0.75 - 0.76) | 0.83 (0.83 - 0.84) | 0.85 (0.84 - 0.86) | 0.86 (0.85 - 0.87) | 0.62 (0.61 - 0.63) | 0.89 (0.88 - 0.89) | 0.83 (0.83 - 0.84) |
| | Bedside SOFA | All | 0.71 (0.71 - 0.72) | 0.77 (0.76 - 0.77) | 0.78 (0.77 - 0.79) | 0.79 (0.78 - 0.79) | 0.59 (0.58 - 0.60) | 0.82 (0.81 - 0.83) | 0.78 (0.77 - 0.79) |
| | | First Only | 0.71 (0.70 - 0.72) | 0.77 (0.76 - 0.78) | 0.79 (0.78 - 0.79) | 0.80 (0.79 - 0.80) | 0.59 (0.58 - 0.60) | 0.83 (0.82 - 0.84) | 0.77 (0.77 - 0.78) |
| | | Unique Only | 0.73 (0.72 - 0.74) | 0.80 (0.79 - 0.80) | 0.81 (0.80 - 0.82) | 0.82 (0.81 - 0.83) | 0.59 (0.58 - 0.60) | 0.85 (0.85 - 0.86) | 0.80 (0.79 - 0.81) |

**Supplementary Table S7.** Externally validated AUC results for three variants of handling patients with multiple ICU stays. Results shown for first hour after ICU admission, at time of ICU discharge, mean across all hours, and for 24, 48, 72, and 96 hours after ICU admission. "All" denotes using all ICU stays available, and is the setting used in the main manuscript. For patients with multiple ICU stays in their EHR, "first only" denotes removing all but their first ICU stay available, while "unique only" removes these patients altogether, ignoring all of their ICU stays. For each setting, new models were trained and evaluated using the modified datasets. 95% confidence intervals shown in parentheses. Modified ICU stay counts: (1) UFHealth: All = 36,216; First Only = 27,660; Unique Only = 22,229; (2) MIMIC: All = 45,748; First Only = 35,993; Unique Only = 28,445.